\pdfoutput=1

\documentclass[10pt,twocolumn,letterpaper]{article}

\usepackage{wacv}              
\usepackage[most]{tcolorbox}

\usepackage{multirow}
\usepackage{graphicx}
\usepackage{subcaption} 
\usepackage{caption}

\usepackage{algorithm}   
\usepackage{algorithmic}   
\usepackage{csquotes}
\usepackage{tabularx} 
\usepackage{array}
\usepackage{makecell}  
\usepackage{tabularx,booktabs,adjustbox}
\usepackage{subcaption} 
\usepackage{placeins}   
\definecolor{wacvblue}{rgb}{0.21,0.49,0.74}
\usepackage[pagebackref,breaklinks,colorlinks,allcolors=wacvblue]{hyperref}

\title{More Than Memory Savings: Zeroth-Order Optimization Mitigates Forgetting in Continual Learning}

\author{Wanhao Yu$^{1}$, Zheng Wang$^{2}$, Shuteng Niu$^{3}$, Sen Lin$^{2}$, Li Yang$^{1\dagger}$\\
$^{1}$Department of Computer Science at University of North Carolina at Charlotte  \\ $^{2}$Department of Computer Science at University of Houston \\
$^{3}$Department of Artificial Intelligence $\&$ Informatics at Mayo Clinic\\
$\{$wyu6, lyang50$\}$@charlotte.edu, $\{$zwang214, slin50$\}$@cougarnet.uh.edu, niu.shuteng@mayo.edu}

\begin{document}
\maketitle

\begingroup
\renewcommand\thefootnote{}\footnotetext{$^\dagger$ Li Yang is the corresponding author.}\addtocounter{footnote}{0}
\endgroup

\begin{abstract}
Zeroth-order (ZO) optimization has gained attention as a memory-efficient alternative to first-order (FO) methods, particularly in settings where gradient computation is expensive or even impractical. Beyond its memory efficiency, in this work, we investigate ZO optimization for continual learning (CL) as a novel approach to address the plasticity-stability-efficiency trilemma.
Through theoretical analysis and empirical evidence, we show that ZO optimization naturally leads to flatter loss landscapes, which in turn reduce forgetting in CL. However, this stability comes at a cost of plasticity: due to its imprecise gradient estimates and slower convergence, ZO optimization tends to be less effective than FO in acquiring new task-specific knowledge, particularly under constrained training budgets.
To better understand this trade-off, we conduct a holistic evaluation of ZO optimization applied to various existing CL methods. Our findings reveal that ZO optimization enhances stability but often undermines plasticity, particularly when used with learnable classifiers.
Motivated by this insight, we propose \textbf{ZO-FC}, a simple but effective approach that applies ZO optimization to a single adapter-based PEFT module with FO optimized classifier. This design leverages the stability benefits of ZO while preserving the adaptability of FO updates with negligible memory overhead. Experiments demonstrate that ZO-FC achieves an effective balance between stability and plasticity, offering a practical and memory-efficient solution for on-device CL. Our code is available at: \href{https://github.com/uncc-efficient-ai/ZOFC}{https://github.com/uncc-efficient-ai/ZOFC}.


\end{abstract}
\section{Introduction}
\label{sec:intro}

Emulating the human ability to learn continuously from new experiences is a key inspiration in artificial intelligence. This pursuit has driven significant attention to Continual Learning (CL) \cite{hadsell2020embracing,  de2021continual, wang2024comprehensive}, a paradigm aiming to train models on a sequence of tasks while preserving previously acquired knowledge. A primary challenge of CL is to maintain a balance between \textit{plasticity}, the ability to learn new knowledge, and \textit{stability}, the ability to retain previously learned knowledge by preventing \textbf{catastrophic forgetting}: a phenomenon where models lose prior knowledge when training on new tasks \cite{FRENCH1999128}.

\label{sec:method}
\begin{figure}[t]
  \centering
  \includegraphics[width=1.0\linewidth]{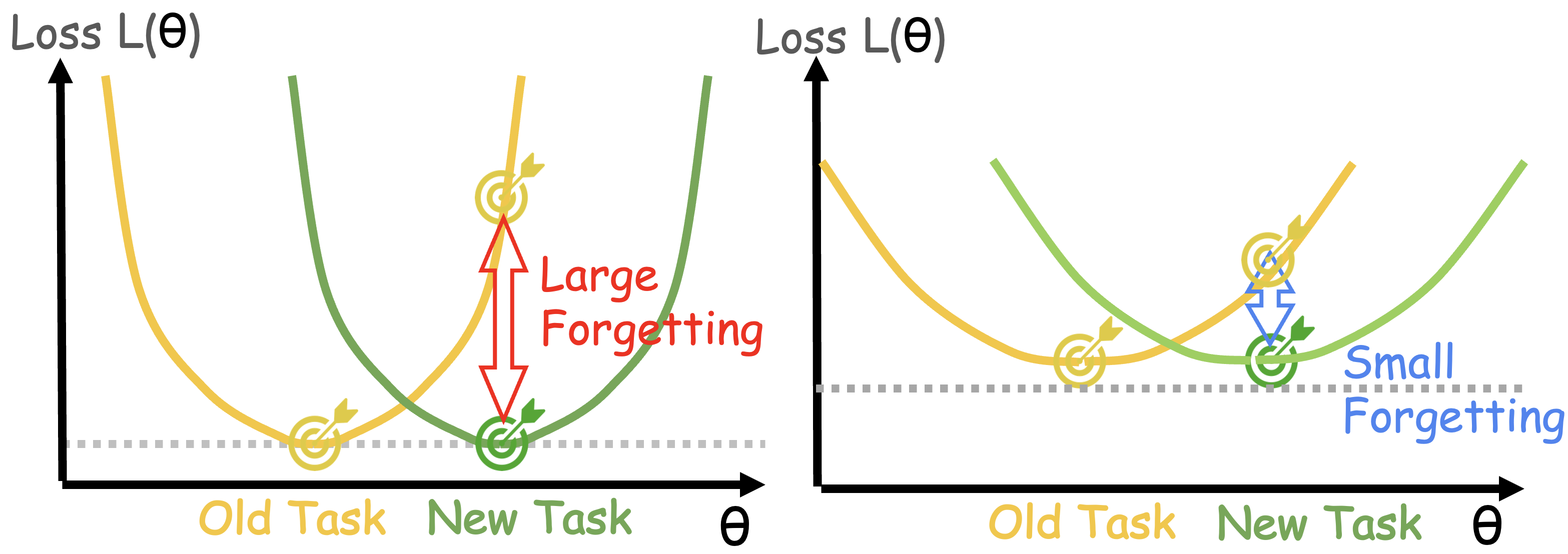}
  \begin{minipage}{0.48\linewidth}
    \centering
    \textbf{(a)}
  \end{minipage}
  \hfill
  \begin{minipage}{0.48\linewidth}
    \centering
    \textbf{(b)}
  \end{minipage}
    \vspace{-0.3em}
  \caption{Illustrating the difference between FO and ZO optimization in addressing the \textit{plasticity–stability} dilemma in continual learning:
(a) FO optimization offers high plasticity, enabling effective learning of new tasks, but suffers from low stability, leading to significant forgetting of previous knowledge;
(b) In contrast, ZO optimization exhibits high stability with reduced forgetting, but lower plasticity, making it less effective at acquiring new task-specific knowledge.}
  \label{fig:loss_concept}
\end{figure}

To address this plasticity-stability dilemma, recent progress in CL has shifted from ``continual training from scratch" to a new paradigm of ``continual fine-tuning from pre-trained models". In particular, driven by the increasingly powerful pre-trained foundation models, such as Vision Transformers (ViTs) \cite{dosovitskiy2021an, steiner2021train}, and the effectiveness of parameter-efficient fine-tuning~(PEFT) techniques (i.e., prompts \cite{lester2021prompt}, adapters \cite{houlsby2019adapter}, or LoRAs \cite{hu2022lora}), recent CL methods \cite{wang2022l2p,gao2023lae,zhou2024aper,zhou2024ease,liang2024inflora} follow the recipe to freeze the backbone and train lightweight modules across tasks. This setup inherits strong plasticity that new tasks are learned quickly from a good initialization so the critical challenge becomes mitigating catastrophic forgetting to enhance stability. These methods therefore isolate task-specific information into PEFT modules to minimize interference with previously learned tasks.


In this work, we take a step back to revisit the plasticity-stability dilemma from the perspective of the optimization process itself. We argue that the \textit{high plasticity low stability} characteristic of PEFT-based CL methods is intrinsically related to standard first-order gradient-based optimization. The precise nature of gradient descent enables the model to quickly reach sharp, low-loss minima, resulting in high accuracy for the current task and strong plasticity. However, these sharp minima are inherently sensitive, meaning that small changes introduced by learning new tasks may disturb the model parameters, causing performance drops on previously learned tasks and thereby undermining stability, as shown in Figure \ref{fig:loss_concept}(a).

Recently, zeroth-order (ZO) optimization has shown promising results in fine-tuning large language models \cite{malladi2023fine, zhang2024revisiting}, providing a memory-efficient alternative to traditional first-order (FO) optimization while achieving comparable accuracy. ZO optimization eliminates the need for backpropagation and the associated storage of intermediate activations, making training memory close to inference.
In this work, we target low-memory, on-device CL settings and further look beyond this established memory advantage. We hypothesize that ZO's gradient-free search makes it naturally tend to wider, flatter minima, which is known to be more robust to catastrophic forgetting~\cite{foret2021sharpnessaware,shi2021fclflat}. 
Motivated by this potential, we investigate ZO optimization for CL as a novel approach to address the \textit{plasticity-stability-efficiency} trilemma, aiming to achieve accuracy comparable to FO methods at a fraction of the training memory overhead.
To this end, our contribution can be summarized as follows:
\begin{itemize}
    \item We provide a formal analysis linking ZO optimization to the plasticity-stability dilemma, showing that its inherent stochastic search promotes convergence to flatter minima, enhancing stability and mitigating forgetting, while also revealing its limitation in plasticity due to imprecise gradient estimation, as shown in Figure \ref {fig:loss_concept}(b).
    \item We conduct a comprehensive empirical investigation and discover that naively applying ZO to existing PEFT-based CL methods fails, specifically due to the learnable classifiers. Based on this insight, we propose \textbf{\textit{ZO-FC}}, a simple and yet effective method that combines the \underline{\textbf{ZO}} optimization PEFT module with the standard \underline{\textbf{f}}irst-order optimization for the \underline{\textbf{c}}lassifier.
    \item Empirically, our proposed ZO-FC overcomes the catastrophic failure of naive ZO substitutions. It achieves accuracy comparable to state-of-the-art FO 
    methods across 3 benchmarks, while using 6x less training memory. Furthermore, we empirically verify that ZO-FC guides the model toward wider, flatter regions, demonstrating the direct link between its optimization path and enhanced stability. We hope our work inspires future research to further explore ZO optimization as a promising direction for tackling the plasticity–stability–efficiency trade-off, especially in resource-constrained CL scenarios.
\end{itemize}

\section{Related Work}
\label{sec:preliminary}


\subsection{PEFT-based Class-incremental Learning}
Class-incremental learning (CIL) is a widely used and challenging setting in continual learning, where new groups of classes are introduced sequentially, and the model must learn to classify all previously seen classes without access to task or class identity during inference~\cite{zhu2021class}. With the growing adoption of large-scale pre-trained foundation models (e.g., Vision Transformer \cite{dosovitskiy2021an, steiner2021train}), a common learning paradigm in CIL is to use the pre-trained model as a strong initialization and perform continual fine-tuning on sequential tasks using parameter-efficient fine-tuning (PEFT), 
while keeping the pre-trained backbone frozen. PEFT enables efficient adaptation by introducing lightweight trainable modules, with common techniques including prompts \cite{lester2021prompt}, adapters \cite{houlsby2019adapter}, and LoRAs \cite{hu2022lora}. These PEFT-based CIL methods \cite{wang2022l2p,gao2023lae,zhou2024aper,zhou2024ease,liang2024inflora} have shown strong performance, even under rehearsal-free settings where no data from previous tasks is used during the training of new tasks. 

By leveraging strong pre-trained models as initialization, PEFT-based CIL methods inherently exhibit high plasticity. As a result, their primary focus is on mitigating forgetting to enhance stability in learning each task. First of all, SimpleCIL~\cite{zhou2024aper} achieves good accuracy, particularly on simpler datasets such as CIFAR100, by only applying prototype-based classification on top of a frozen pre-trained model without any learning. This result empirically highlights the effectiveness of leveraging pre-trained representations. Moreover, to mitigate forgetting when introducing task adaptive PEFT modules, a series of methods has been proposed. 
L2P \cite{wang2022l2p} uses a learnable pool of prompts to activate task-specific representations, effectively isolating knowledge across tasks. CODA-Prompt \cite{smith2023codap} refines L2P by applying contrastive alignment to ensure consistent representation. LAE \cite{gao2023lae} maintains an online PEFT module for current-task learning and an offline PEFT module updated via exponential moving average (EMA) so that these modules can be ensembled at inference to combine plastic adaptation and stable accumulation. APER~\cite{zhou2024aper} extends SimpleCIL to adapt the pre-trained model only during the first task, then merges features from both the frozen pre-trained backbone and the adapted model to form a merged feature extractor. EASE \cite{zhou2024ease} adopts a task-specific adapter for each task, and introduces a semantic-guided prototype synthesis strategy that reconstructs old class representations in the current subspace to mitigate forgetting. InfLoRA \cite{liang2024inflora} adapts LoRA-based CL with an interference-free design by constraining the update directions of LoRA modules to be task-orthogonal to reduce cross-task interference. In this work, these methods serve as FO optimization baselines for evaluating the application of ZO optimization.

\subsection{Zeroth-Order Optimization}
Zeroth-order (ZO) optimization has been extensively studied \cite{spall1987spsa, nesterov2017gradientfree, liu2018zeroth, liu2020primer} as a gradient-free alternative to first-order (FO) methods. By estimating gradients using only function evaluations without the need for backpropagation, ZO optimization is appealing in privacy-sensitive, black-box, or resource-constrained settings, such as federated learning and edge device deployment \cite{chen2017zoo, tu2019autozoom, fang2022communication, shu2023federated}. A representative ZO method is Simultaneous Perturbation Stochastic Approximation (SPSA)~\cite{spall1987spsa}, which estimates gradients by computing finite differences of the loss function along randomly sampled direction vectors. Given model parameters $\theta$ and a loss function $L$, the ZO gradient estimate can be expressed as:
\begin{equation}
\hat{g} = \frac{L(\theta + \varepsilon \Delta) - L(\theta - \varepsilon \Delta)}{2\varepsilon} \cdot \Delta^{-1},
\label{eq:spsa}
\end{equation}
where $\Delta \sim {\pm 1}^d$ is a random perturbation vector, $d$ is the parameter dimension, and $\varepsilon$ is a small perturbation scale. This ZO estimate can be further incorporated into standard optimizers, such as stochastic gradient descent (SGD), to yield batch-wise zeroth-order stochastic gradient descent (ZO-SGD)~\cite{ghadimi2013stochastic, nesterov2017gradientfree}.  To reduce the high variance inherent in single-direction estimates for each updating step, a common practical extension is to average over $Q$ queries of independent perturbation directions per gradient step \cite{spall2000adaptive}. 

Notably, recent work MeZO \cite{malladi2023fine} has demonstrated that ZO-SGD fine-tuning on large language models can achieve accuracy comparable to its FO counterpart. In addition, while there are other ZO techniques, including Forward-Grad~\cite{baydin2022gradients, ren2022scaling}, Sign-based variants~\cite{liu2019signsgd, malladi2023fine}, and adaptive methods such as ZO-AdaMM~\cite{chen2019zo}, recent benchmark studies~\cite{zhang2024revisiting, feng2025zeroflow} show that no single method consistently outperforms others in all tasks. Therefore, we adopt SPSA and its mini-batch extension, ZO-SGD, as our default ZO optimization strategy.

\section{Rethinking Zeroth-Order Optimization for Continual Learning}
\label{sec:theory}

In this work, we intuitively hypothesize that ZO optimization is inherently well-suited for mitigating catastrophic forgetting, thereby enhancing the stability of continual learning (CL) when adapting to new tasks. To support this hypothesis, we provide a theoretical analysis that connects ZO optimization to the core challenge of \textit{plasticity-stability} dilemma in CL. Our analysis is grounded in two key lines of research: (1) flat minima of loss landscape are known to reduce forgetting, and (2) ZO optimization tends to favor convergence to flatter minima.


\noindent\textbf{1) Loss Flatness Mitigates Forgetting: }
Let $\theta_t\!\in\!\mathbb{R}^{d}$ denote the model parameters after training on task~$t$. When learning 
task~$t{+}1$, the parameters are updated by  $\Delta\theta=\theta_{t+1}-\theta_{t}$.
Following~\cite{mirzadeh2021linear},
we quantify \textit{forgetting} on previous tasks as the increase in their cumulative loss:
\begin{equation}
F_t =
L_{\text{old}}\!\bigl(\theta_t+\Delta\theta\bigr)
-\,L_{\text{old}}\!\bigl(\theta_t\bigr),
\label{eq:forget-def}
\end{equation}
\noindent where $L_{\text{old}}(\theta) = \sum_{i=1}^t L_i(\theta)$ represents the sum of per-task losses of all previous tasks. A second-order Taylor expansion of $L_{\text{old}}$ around $\theta_t$ gives:
\begin{equation}
F_t =
\tfrac12\,\Delta\theta^{\!\top} H_t\,\Delta\theta
+O(\|\Delta\theta\|^{3})
\label{eq:forget-taylor}
\end{equation}
where $H_t =\nabla^{2}L_{\text{old}}(\theta_t)$, representing the Hessian of the old-task loss. As proved by prior works~\cite{mirzadeh2020understanding}, the forgetting can be bounded by:
\begin{equation}
F_t \;\le\;
\tfrac12\,\lambda_{\max}(H_t)\,
\|\Delta\theta\|^{2}
\label{eq:forget-bound}
\end{equation}
where $\lambda_{\max}(H_t)$ is the largest eigenvalue of the Hessian matrix at task $t$. As revealed by~\cite{keskar2017largebatch}, 
this eigenvalue captures the sharpness of the loss landscape: a higher $\lambda_{\max}(H_t)$ indicates a sharper curvature, meaning that small changes in parameters $\Delta\theta$ can result in large increases in loss. As a result, models operating in sharper regions are more prone to catastrophic forgetting, since even minor updates can severely disrupt performance on previously learned tasks.

Furthermore, recent work~\cite{kong2024hessian} empirically confirms that smaller curvature (i.e., lower $\lambda_{\max}(H_t)$) corresponds to flatter regions of the loss surface, which are associated with reduced forgetting in continual learning. Motivated by this, several studies~\cite{mirzadeh2021linear, wu2024cr, bian2024make, shi2021fclflat} have proposed augmenting standard FO optimization with auxiliary loss terms or steps aimed at minimizing curvature surrogates to promote flatter minima.
In contrast, we aim to explore mitigating forgetting through the intrinsic bias of ZO optimization.

\noindent\textbf{2) ZO optimization tends to flatter minima: }
Foundational work \cite{nesterov2017gradientfree} has theoretically shown that 
every ZO optimization step (Eq.\ref{eq:spsa}) is, in fact, an update on a smoothed version of the original loss. For any twice-differentiable loss $L(\theta)$ and an isotropic random direction $\Delta$, the smooth loss is defined as the following expected loss:
\begin{equation}
L_{\varepsilon}(\theta) = \mathbb{E}_{\Delta}[L(\theta + \varepsilon \Delta)].
\label{eq:smooth-loss}
\end{equation}
A second-order Taylor expansion gives:
\begin{equation}
L_{\varepsilon}(\theta) = L(\theta) + \frac{\varepsilon^2}{2} \text{Tr}(\nabla^2 L(\theta)) + O(\varepsilon^4),
\label{eq:smooth-loss-expansion}
\end{equation}
where the noise term $O(\varepsilon^4)$ is negligible. The extra term $\frac{\varepsilon^2}{2} \operatorname{Tr}(\nabla^2 L)$ is proportional to the trace of the Hessian, which can be seen as a measure of curvature of the loss landscape \cite{zhang2025zofindsflat}. Minimizing $L_\varepsilon$ through ZO optimization therefore minimizes the original loss and simultaneously penalizes large positive curvature. Because sharp minima possess large positive Hessian trace, they receive a larger penalty in $L_\varepsilon$. ZO SPSA updates, which descend on this smoothed objective, are statistically steered away from narrow valleys and into flatter regions of the landscape. When ZO optimization converges to a stationary point of $L_\varepsilon$, the trace of the Hessian of the original loss at that point is also minimized. \textbf{Therefore, ZO optimization inherently favors flat minima without the need for explicit regularization.}


\noindent\textbf{The untapped benefit and limitation of ZO optimization for \textit{plasticity-stability} dilemma: }
Given the above analysis that reveals 1) flatter loss landscapes help mitigate forgetting by reducing sensitivity to parameter changes and 2) ZO optimization naturally tends toward flatter minima, we can build a connection that:
\begin{equation}
\text{ZO optimization} \Longrightarrow 
\lambda_{\max}(H_t)\!\downarrow 
\Longrightarrow F_t\downarrow\
\label{eq:flat-to-forget}
\end{equation}
To this end, it suggests that ZO optimization naturally obtains \textbf{\textit{high stability}} in continual learning by mitigating catastrophic forgetting ($F_t$).

\noindent However, this benefit comes with a trade-off in \textit{plasticity}.  Unlike FO optimization, which efficiently converges toward local minima via accurate gradient information, ZO optimization relies on stochastic perturbations and typically exhibits slower convergence. Specifically, the convergence rate of ZO methods is $O(d/Q)$, worse than their FO counterparts \cite{ghadimi2013stochastic}, where $d$ is the parameter dimension and $Q$ is the number of random queries per step, making them less efficient than FO \cite{nesterov2017gradientfree}. As a result, given the limited training budget for CL, this slow convergence may prevent ZO methods from reaching sufficiently good local minima, thus impairing \textit{plasticity} and overall task adaptation.

\begin{figure*}[h]
  \centering
  \vspace{-1.0em}
  \includegraphics[width=0.28\linewidth]{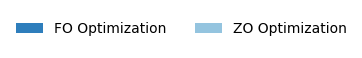}
  \vspace{-0.5em}

  \begin{subfigure}[t]{0.48\textwidth}
    \centering
    \includegraphics[width=0.92\linewidth]{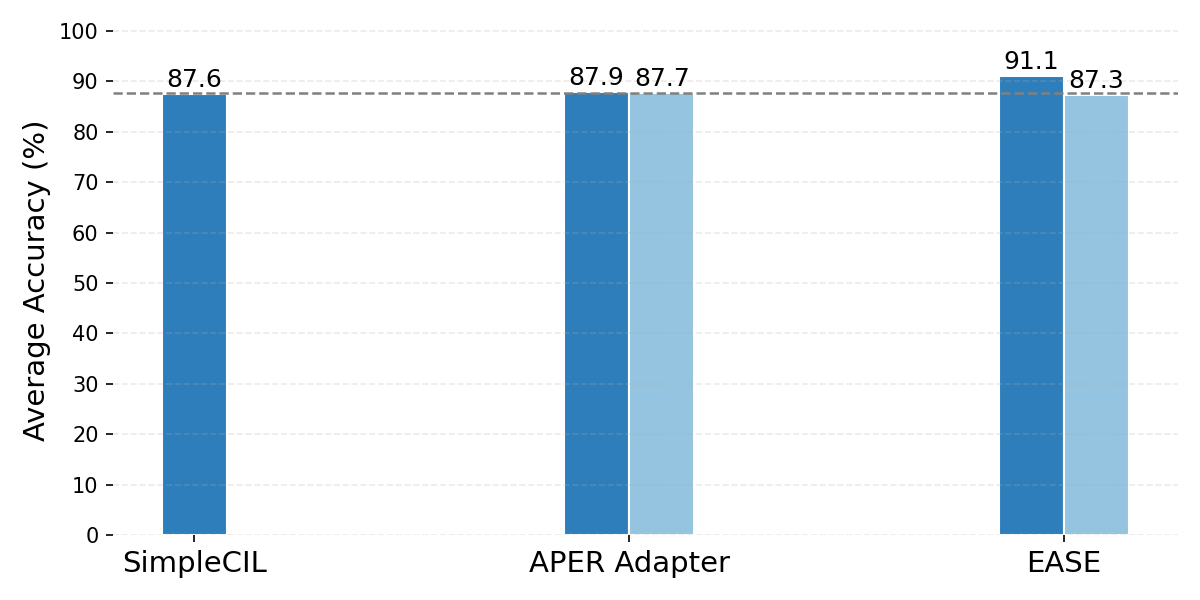}
    \caption{Methods with prototype classifiers on CIFAR100.}
    \label{fig:cifar100_proto}
  \end{subfigure}
  \hfill
  \begin{subfigure}[t]{0.48\textwidth}
    \centering
    \includegraphics[width=0.92\linewidth]{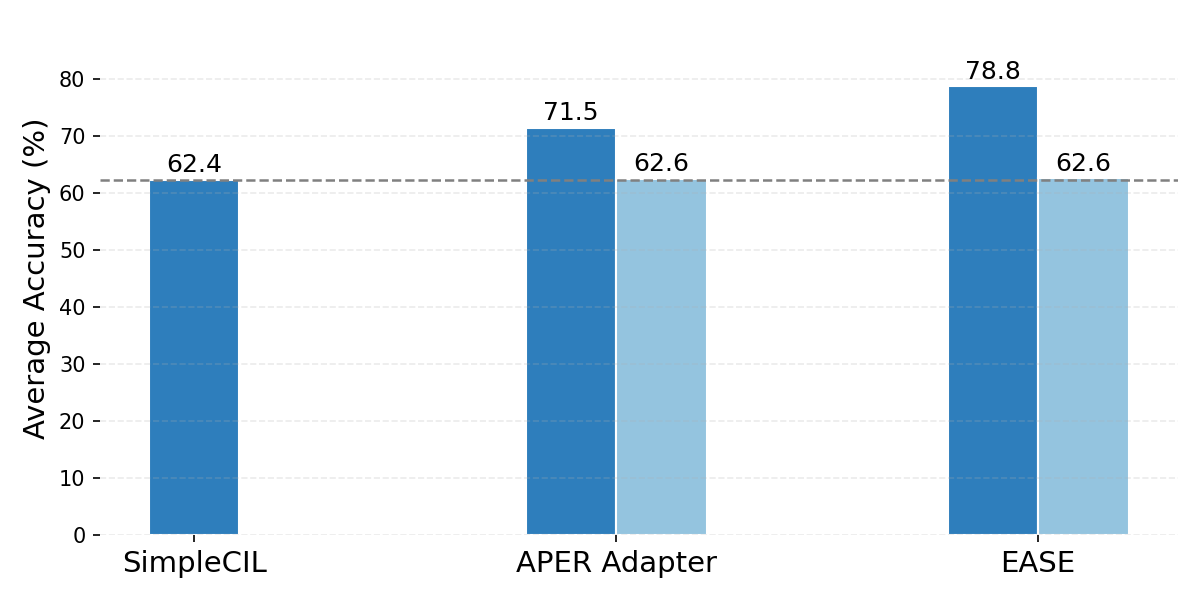}
    \caption{Methods with prototype classifiers on ImageNet-R.}
    \label{fig:inr_proto}
  \end{subfigure}

  \vspace{-0.1em}  

  \begin{subfigure}[t]{0.48\textwidth}
    \centering
    \includegraphics[width=0.92\linewidth]{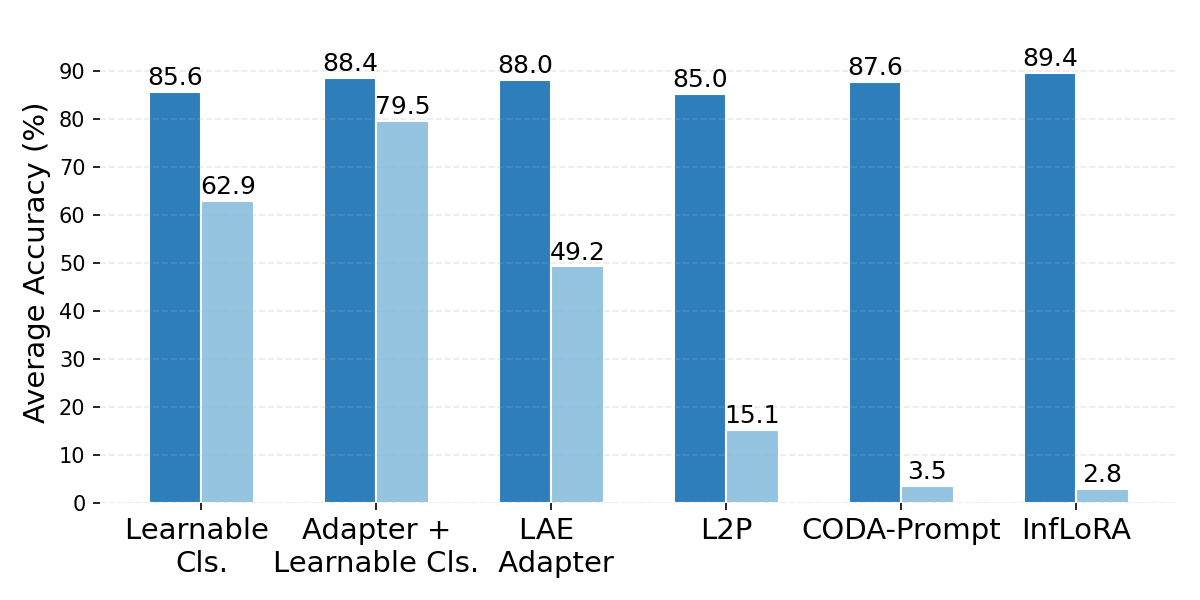}
    \caption{Methods with learnable classifiers on CIFAR100.}
    \label{fig:cifar100_cls}
  \end{subfigure}
  \hfill
  \begin{subfigure}[t]{0.48\textwidth}
    \centering
    \includegraphics[width=0.92\linewidth]{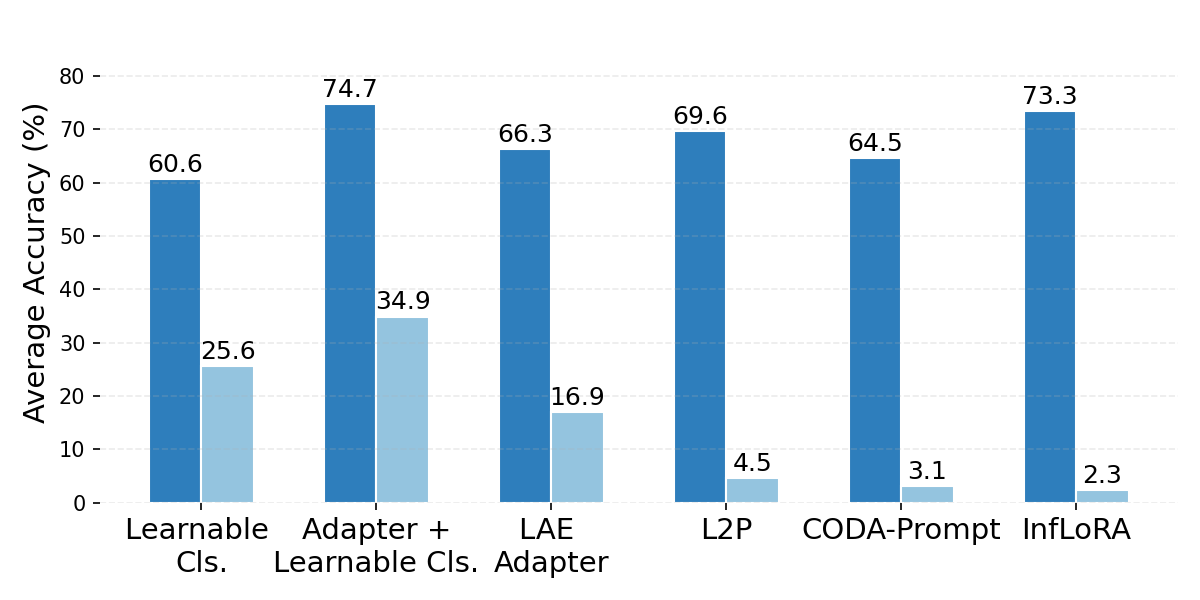}
    \caption{Methods with learnable classifiers on ImageNet-R.}
    \label{fig:inr_cls}
  \end{subfigure}

  \caption{Comparison of FO and ZO optimization methods under 5-class incremental learning on CIFAR100 and ImageNet-R.
  }
  \label{fig:fo_zo_comparison}
\end{figure*}

\section{Method}
\label{sec:method}
\subsection{Direct ZO Substitution in CL Fails}
\label{sec:zo_failure}

To investigate the effectiveness of ZO optimization in CL, we begin with a comprehensive empirical evaluation by applying ZO optimization directly to existing PEFT-based methods. Specifically, we select 8 representative approaches, whose FO baseline performance on CIFAR100 and ImageNet-R is summarized in Table \ref{tab:main_results}. Moreover, as shown in Table \ref{tab:clf_families}, these methods can be broadly categorized based on their classifier types: (1) prototype-based classifiers, which compute predictions by measuring the distance between input features and task-specific class prototypes; and (2) unified learnable classifiers, which maintain a single trainable classification head shared across all observed tasks.
To ensure a fair comparison, we use a consistent pre-trained ViT backbone and standardized training configurations across all experiments. 
It is important to note that due to the finite-difference nature of ZO optimization (as shown in Eq.~\ref{eq:spsa}), each ZO update requires two forward passes per iteration.
From the results shown in Figure \ref{fig:fo_zo_comparison},
we find that \textbf{naively replacing FO optimization with ZO optimization leads to consistent failure across all evaluated methods}. Interestingly, the two classifier families exhibit notably distinct behaviors. Such that, to better understand the discrepancy, we further analyze the results of each classifier family separately in the following sections.




\begin{table}[t]
\footnotesize
\setlength{\tabcolsep}{3pt}
\renewcommand{\arraystretch}{1.1}

\newcolumntype{L}{>{\raggedright\arraybackslash}p{0.24\columnwidth}}
\newcolumntype{C}{>{\raggedright\arraybackslash}p{0.47\columnwidth}}
\newcolumntype{R}{>{\raggedright\arraybackslash}p{0.28\columnwidth}}

\begin{tabularx}{\columnwidth}{@{}L C R@{}}
\toprule
\textbf{Classifier family} & \textbf{Decision Mechanism} & \textbf{ Methods}\\
\midrule
\makecell[l]{\textbf{Prototype}\cite{Rebuffi2016iCaRL}\\(nearest-centroid)} &
\makecell[l]{
Centroid: $c_y=\dfrac{1}{|D_y|}\sum\! f_\theta(x)$\\
Predict: $\hat y=\arg\min_y\|h-c_y\|^{2}$} &
\makecell[l]{SimpleCIL\cite{zhou2024aper}\\
EASE\cite{zhou2024ease}\\
APER\_Adapter\cite{zhou2024aper}} \\[4pt]
\midrule
\makecell[l]{\textbf{Unified learnable}\\(linear / cosine)} &
\makecell[l]{
Linear: $s_y(h)=w_y^{\!\top}h$ \\
Cosine\cite{wang2018cosface}: $s_y(h)=\alpha\dfrac{w_y^{\!\top}h}{\|w_y\|\|h\|}$ \\
Predict: $\hat y=\arg\max_y s_y(h)$} &
\makecell[l]{Adapter\cite{chen2022adapter}\\
LAE\cite{gao2023lae}\\
L2P\cite{wang2022l2p}\\
CODA-Prompt\cite{smith2023codap}\\
InfLoRA\cite{liang2024inflora} }\\
\bottomrule
\end{tabularx}
\caption{Two classifier families used by rehearsal-free PEFT CIL.}
\label{tab:clf_families}
\end{table}

\noindent \textbf{Methods with prototype-based classifier: } 
Among these three methods as shown in Table \ref{tab:clf_families}, SimpleCIL serves as a naive baseline that performs prototype classification without any additional trainable parameters, providing a lower-bound reference for performance.
At first glance, the results in Figure \ref{fig:fo_zo_comparison} (a) and (b) show that applying ZO optimization to EASE and APER seems to yield performance comparable to FO optimization on the CIFAR100 dataset. But this trend does not hold on the more challenging ImageNet-R dataset, where ZO optimization causes significant performance degradation for APER and EASE.

To further interpret this behavior, we compare the ZO-optimized versions of all three methods to the FO baseline of SimpleCIL. It's interesting to find that: 
\textbf{applying ZO optimization to all three methods results in performance that closely matches that of SimpleCIL}, which is the simplest baseline that has no training parameters and only generates prototypes from the frozen backbone of each task for classification.
This observation suggests that applying ZO optimization to the PEFT components in EASE and APER has minimal impact on the decisions made by the prototype-based classifier. We hypothesize that when ZO optimization is applied to PEFT modules, its reliance on random perturbations for gradient estimation may lead to insufficiently informative updates. Without strong directional signals, the optimizer struggles to induce the fine-grained adaptations necessary to shift the feature embeddings meaningfully. As a result, the nearest prototypes remain largely unchanged, and the overall classification behavior is unaffected.


\noindent\textbf{Methods with unified learnable classifier: } 
As shown in Figure \ref{fig:fo_zo_comparison} (c) and (d), ZO optimization leads to a clear accuracy degradation across all methods compared to their FO counterparts. Notably, the extent of degradation varies significantly, with L2P, CODA-Prompt, InfLoRA performing even worse than random guessing on the CIFAR100 dataset.

To better understand these failures, we further examine the training loss and identify distinct underlying causes:
\begin{enumerate}
    \item \textbf{L2P, CODA-Prompt, and InfLoRA.} The main reason ZO variants of these methods perform poorly is that they fail to converge within the unified training budget. These methods depend on gradient-precise regularizers, e.g., prompt routing with auxiliary penalties, contrastive alignment losses and orthogonality constraints. Such objectives rely on low-variance, directionally accurate gradients. SPSA ZO estimators provide only noisy information. With finite query budgets such noise destabilizes those regularizers, leading to divergence. Achieving better performance may therefore require an unaffordable increase in training budget or more advanced ZO gradient estimation methods.
    \item \textbf{Single-adapter and classifier-only methods:} For methods such as ``Adapter + Learnable Classifier", ``LAE Adapter", and ``Learnable Classifier" alone, we observe that the training loss converges far better than the prior 3 methods under ZO optimization. It motivates us to use the single shared PEFT module across tasks for ZO optimization. However, ZO optimized loss remains substantially higher than FO optimized loss, indicating suboptimal solutions. This suggests that while ZO optimization is capable of reducing the loss, its noisy gradient estimates hinder \textit{plasticity}, limiting the model's ability to effectively acquire new knowledge in continual learning.
\end{enumerate}

\vspace{-3pt}
\noindent\textbf{Pitfall of learnable classifier for ZO optimization: } 
In PEFT-based CL methods with a learnable classifier, there are typically two ZO-optimized components: the PEFT module and the classifier. To better understand the impact of each component, particularly in adapter-based methods where training loss converges, we analyze their individual contributions. As shown in Figure \ref{fig:fo_zo_comparison} (c) and (d), we compare ``Adapter + Learnable Classifier" with ``Learnable Classifier", where the only difference is that ``Adapter + Learnable Classifier" further trains an adapter continually upon the frozen backbone. Surprisingly, adding a single adapter optimized via ZO significantly improves accuracy compared to using the learnable classifier alone.
This result suggests that \textbf{ZO optimization applied to the adapter contributes positively by enhancing plasticity, while ZO optimization applied to the classifier is likely the primary source of performance degradation.}
We hypothesize that effective classifier training requires precise, directional updates to establish robust decision boundaries, which ZO optimization, with its stochastic and less accurate gradient estimates, struggles to provide.

\subsection{ZO-FC: ZO optimization for PEFT module with FO learnable classifier}
\begin{figure}[!tbp]
  \centering
  \includegraphics[width=\linewidth]{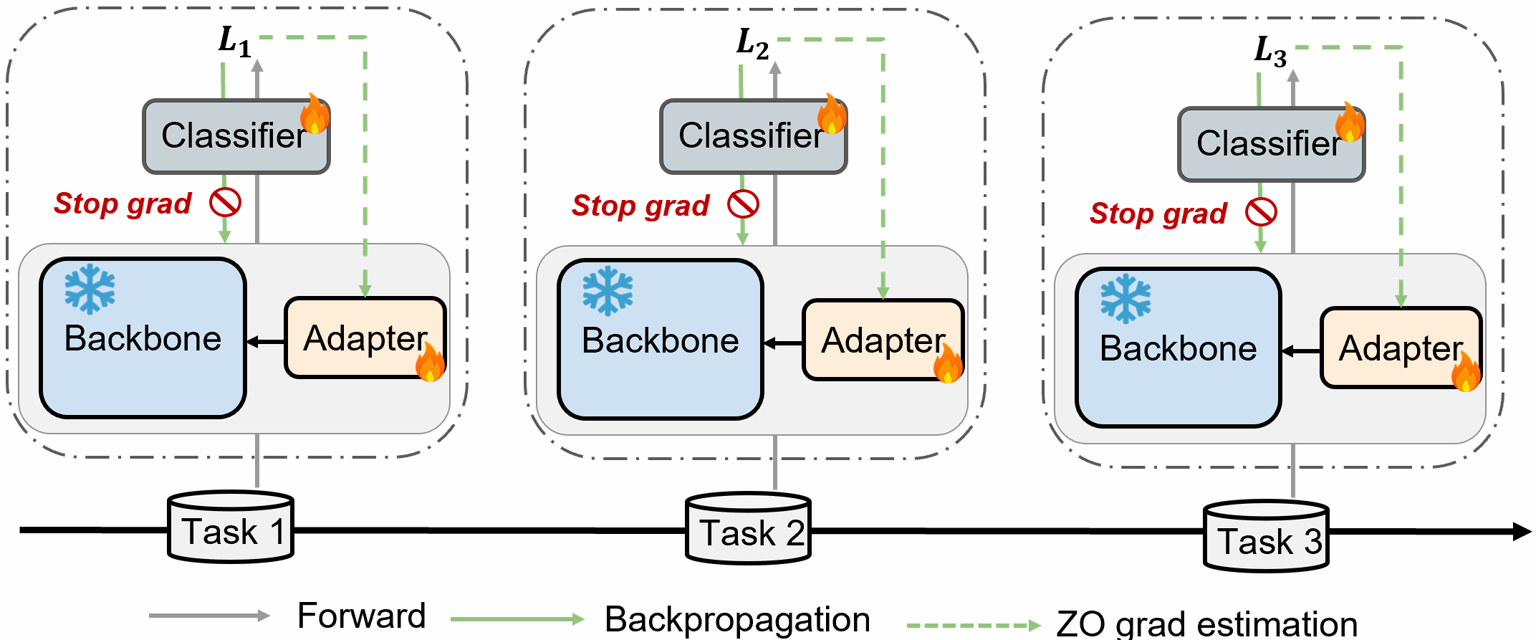}
  \caption{The overview of the proposed ZO-FC.}
  \label{fig:diagram}
\end{figure}

Based on our analysis in Section \ref{sec:zo_failure}, we propose a simple yet effective solution, ZO-FC, which applies ZO optimization to a single PEFT module on top of a frozen pre-trained backbone, while retaining FO optimization for the learnable classifier, as shown in Figure \ref{fig:diagram}. It is important to note that our use of a single-PEFT continual learning setup is motivated by the observation that, when built upon a strong pre-trained backbone, a single PEFT module updated across all tasks serves as a simple yet strong FO baseline as shown in Table \ref{tab:main_results}. In contrast, more complex CL methods, such as those relying on prototype-based classifiers or additional regularization, tend to be less compatible with ZO optimization.
Moreover, through empirical comparison of different PEFT choices, including prompts, adapters, and LoRA, we find that adapters consistently yield the best performance. Therefore, we adopt the adapter as the default PEFT module in ZO-FC. 

\label{sec:flatness_metric} 

Although ZO-FC employs FO optimization for the classifier, it still benefits from the flat minima effect: the ZO updates bias the adapter representation toward flatter solutions, while the FO classifier preserves decision-boundary plasticity (proof in supplementary material). Besides, its impact on memory efficiency remains negligible. This is because the classifier is the final layer in the model and does not propagate gradients back to earlier layers. As a result, intermediate activations from the backbone and adapter do not need to be stored during training, significantly reducing memory usage. Therefore, ZO-FC retains the key advantage of memory efficiency while benefiting from the precision of first-order updates where they matter most. The full procedure is detailed in Algorithm 1
in supplementary material.

\section{Experiments}
\label{sec:exp}

\begin{table*}[h]
  \centering
  \normalsize                     
  \setlength{\tabcolsep}{10pt}             
  \renewcommand{\arraystretch}{1.05}      
  \begin{tabular}{l|ccc|ccc|ccc}
    \hline
    \multicolumn{1}{c|}{\multirow{2}{*}{Method}} &
      \multicolumn{3}{c|}{\textbf{CIFAR100 Inc-5}} &
      \multicolumn{3}{c|}{\textbf{ImageNet-R Inc-5}} &
      \multicolumn{3}{c}{\textbf{DomainNet Inc-69}} \\
      & avg$\uparrow$ & last$\uparrow$ & fgt$\downarrow$
      & avg$\uparrow$ & last$\uparrow$ & fgt$\downarrow$
      & avg$\uparrow$ & last$\uparrow$ & fgt$\downarrow$ \\
    \hline
    \multicolumn{10}{c}{\rule{0pt}{2.5ex}\textbf{Classifier Only}\rule[-1.2ex]{0pt}{0pt}} \\
    \hline\hline
    SimpleCIL\cite{zhou2024aper}        & 87.57 & 81.26 & 6.11 & 62.39 & 54.33 & 8.36 & 65.76 & 59.91 & 7.02 \\
    \hline
    \hline
    \multicolumn{10}{c}{\rule{0pt}{2.5ex}\textbf{FO Optimization}\rule[-1.2ex]{0pt}{0pt}}\\
    \hline\hline
    Learnable Cls.\cite{Hou2019LUCIR}   & 85.61 & 78.33 & \textbf{5.17} & 60.60 & 52.12 & \textbf{3.98} & 75.85 & 69.88 & \textbf{3.70} \\
    Adapter + Cls.                      & 88.41 & 83.64 & 9.58 & 74.71 & 69.62 & 8.49 & 78.67 & 72.85 & 4.64 \\
    LAE Adapter\cite{gao2023lae}        & 88.05 & 82.31 & 8.35 & 66.28 & 55.78 & 9.93 & 74.55 & 68.03 & 15.17 \\
    L2P\cite{wang2022l2p}               & 85.05 & 78.59 & 9.09 & 69.60 & 62.35 & 6.12 & 76.39 & 70.30 & 12.22 \\
    CODA-Prompt\cite{smith2023codap}    & 87.58 & 80.60 & 7.04 & 64.54 & 55.23 & 5.30 & 78.84 & 73.29 & 10.20 \\
    APER Adapter\cite{zhou2024aper}     & 87.90 & 81.68 & 6.01 & 71.52 & 63.42 & 7.48 & 71.24 & 64.93 & 6.72 \\
    EASE\cite{zhou2024ease}             & \textbf{91.13} & \textbf{85.45} & 7.05 & \textbf{78.78} & \textbf{71.93} & 7.54 & 75.91 & 70.24 & 8.54 \\
    InfLoRA\cite{liang2024inflora}      & 89.41 & 82.89 & 6.95 & 73.33 & 64.52 & 7.95 & \textbf{79.70} & \textbf{74.05} & 7.43 \\
    \hline
    \hline
    \multicolumn{10}{c}{\rule{0pt}{2.5ex}\textbf{ZO Optimization}\rule[-1.2ex]{0pt}{0pt}}\\
    \hline\hline
    Learnable Cls.                      & 62.85 & 43.55 & 3.99 & 25.60 & 17.08 & 9.18 & 52.70 & 46.21 & 7.37 \\
    Adapter + Cls.                      & 79.45 & 70.46 & 8.72 & 34.87 & 27.33 & 6.36 & 48.24 & 44.29 & 4.56 \\
    LAE Adapter                         & 49.18 & 36.49 & 18.43 & 16.92 & 8.87 & 8.23 & 34.62 & 26.94 & 8.54 \\
    L2P                                 & 15.05 &  2.36 & 14.02 &  4.55 &  1.42 & 4.24 & 17.89 & 14.78 & 11.86 \\
    CODA-Prompt                         &  3.53 &  1.21 &  2.49 &  3.10 &  0.92 & 1.90 &  3.09 &  1.12 &  2.17 \\
    APER Adapter                        & 87.70 & 81.49 &  6.07 & 62.60 & 54.50 & 8.55 & 62.45 & 56.67 &  6.89 \\
    EASE                                & 87.32 & 81.07 &  6.87 & 62.65 & 54.47 & 9.84 & 63.10 & 56.80 &  7.22 \\
    InfLoRA                             &  2.78 &  0.94 &  2.29 &  2.26 &  0.89 & 1.97 &  4.78 &  3.23 &  2.65 \\
    \hline
    \hline
    \textbf{ZO-FC (Ours)}                        & \textbf{88.39} & \textbf{83.34} & \textbf{5.26} & \textbf{72.01} & \textbf{66.63} & \textbf{4.40} & \textbf{77.17} & \textbf{71.05} & \textbf{4.19} \\
    \hline
  \end{tabular}
  \caption{Results(\%) on CIFAR100, ImageNet-R and DomainNet with existing FO PEFT-based CIL methods and their ZO counterparts.}
  \label{tab:main_results}
\end{table*}

\subsection{Experimental Details}
\label{sec:setup}

We evaluate the methods on three widely-used benchmarks in CIL works: CIFAR100 \cite{krizhevsky2009learning}, ImageNet-R \cite{russakovsky2015imagenet} and DomainNet \cite{peng2019moment}. Following the settings in prior work \cite{rebuffi2017icarl, smith2023codap}, we conduct the experiments using two settings: 5-class and 10-class per task for CIFAR100 and ImageNet-R, 69-class per task for DomainNet. We fix the class/task order using seed 1993 for all methods and datasets.

\noindent\textbf{Evaluation Metric: }
Across all experiments, we report three key metrics following \cite{sun2025pilot}: (1) \textbf{average accuracy (avg)} across all tasks at the end of training, (2) \textbf{last-task accuracy (last)} on the latest task, and (3) \textbf{catastrophic forgetting (fgt)}, defined as the average drop in accuracy of old tasks from their best observed accuracy to their final accuracy \cite{díazrodríguez2018dontforgetforgettingnew}. Formal definitions are provided in supplementary material.

\noindent\textbf{Training Details: }
All methods use a ViT-B/16~\cite{dosovitskiy2021an} backbone pre-trained on ImageNet-21K \cite{russakovsky2015imagenet}. Following~\cite{gao2023lae}, we insert a LoRA-like adapter module with rank $r=5$ in parallel to the MLP layers in each of the first five transformer blocks. For ZO optimization, we set the perturbation magnitude to $\varepsilon=0.001$ and the query budget to $Q=4$ with L2 clipping at 1.0 for the ZO updates. In ZO-FC, the classifier is optimized with SGD (LR=0.01, cosine decay), and the adapter with ZO-SGD (LR=0.01, constant). Note that, we adopt the official implementations and follow the hyperparameter settings recommended in their original papers and the LAMDA-PILOT benchmark \cite{sun2025pilot}. And the compared methods are conducted by using same checkpoints of the pre-trained model and task orders to ensure a fair comparison. The other related training configurations can be found in supplementary material.

\subsection{Main Results}
Table \ref{tab:main_results} illustrates the performance comparison with existing CIL methods and their ZO optimization counterparts. 
First of all, compared to the other ZO optimization that applied to the PEFT-based methods, our method consistently achieves significantly better accuracy and low forgetting. For example, compared to ZO-EASE, ZO-FC boosts accuracy over 12$\%$ last accuracy on ImageNet-R dataset. In addition, compared to original FO PEFT-based CIL baselines, the proposed ZO-FC achieves better last accuracy compared to 6 prior methods on CIFAR100 and ImageNet-R, and only behind ``Adapter + Cls." and EASE methods. Furthermore, to evaluate robustness under domain shift, we conduct experiments on DomainNet \cite{peng2019moment}. ZO-FC consistently achieves the lowest forgetting among FO PEFT-based CIL baselines except the learnable classifier, while also outperforming 5 prior methods and achieving a clear performance gain over their ZO variants. In addition, under the 10-incremental setup on CIFAR100 and ImageNet-R in Table \ref{tab:inc10_results}, the proposed ZO-FC achieves the lowest forgetting while maintaining competitive accuracy compared to the FO baselines.

\begin{table}[h]
  \centering
  \small
  \setlength{\tabcolsep}{3pt}  
  \begin{tabular}{lcccccc}
    \toprule
    \multirow{2}{*}{Method} &
      \multicolumn{3}{c}{\textbf{CIFAR100 Inc-10}} &
      \multicolumn{3}{c}{\textbf{ImageNet-R Inc-10}} \\
    \cmidrule(lr){2-4}\cmidrule(lr){5-7}
      & avg$\uparrow$ & last$\uparrow$ & fgt$\downarrow$ & avg$\uparrow$ & last$\uparrow$ & fgt$\downarrow$ \\
    \midrule
    SimpleCIL & 87.13 & 81.26 & 6.08 & 61.99 & 54.55 & 8.08 \\
    \midrule
    \multicolumn{7}{c}{\textbf{FO Optimization}}\\
    \midrule
    \midrule
    Learnable Cls. & 87.03 & 80.92 & 4.31 & 65.67 & 57.52 & \textbf{3.67} \\
    Adapter + Cls. & 92.04 & \textbf{88.54} & 5.48 & 78.89 & 73.27 & 7.04 \\
    LAE Adapter    & 91.31 & 86.54 & 5.58 & 71.68 & 62.43 & 7.95 \\
    L2P            & 87.59 & 82.43 & 11.56 & 74.14 & 68.08 & 6.45  \\
    CODA-Prompt     & 90.76 & 85.7 & \textbf{4.23} & 74.15 & 67.22 & 4.56  \\
    APER Adapter  & 92.15 & 87.41 & 4.42 & 73.20 & 65.88 & 7.04  \\
    EASE           & \textbf{92.10} & 87.67 & 5.94 & \textbf{80.64} & \textbf{74.38} & 7.41  \\
    InfLoRA       & 91.60 & 86.44 & 4.60 & 79.54 & 71.08 & 9.01 \\
    \midrule
    \midrule
    \textbf{ZO-FC (Ours)}  & \textbf{91.13} & \textbf{86.93} & \textbf{3.42}   & \textbf{75.80} & \textbf{70.10} & \textbf{3.64} \\
    \bottomrule
  \end{tabular}
  \caption{Results(\%) on CIFAR100 and ImageNet-R under Inc-10 setup.}
  \label{tab:inc10_results}
\end{table}



\noindent\textbf{ZO-FC exhibits strong stability for continual learning: } 
By diving into the results in Table \ref{tab:main_results} and Table \ref{tab:inc10_results}, ZO-FC consistently has lower forgetting than other PEFT FO baselines, and in Inc-10 it is even below Learnable Cls., which updates only the classifier with FO. This indicates that \textbf{the primary source of forgetting in ZO-FC arises from the first-order optimized classifier, while the ZO-optimized adapter remains highly stable and exhibits minimal additional forgetting combined with the classifier.}

\begin{figure}[h]              
  \centering
  \includegraphics[width=1.0\linewidth]{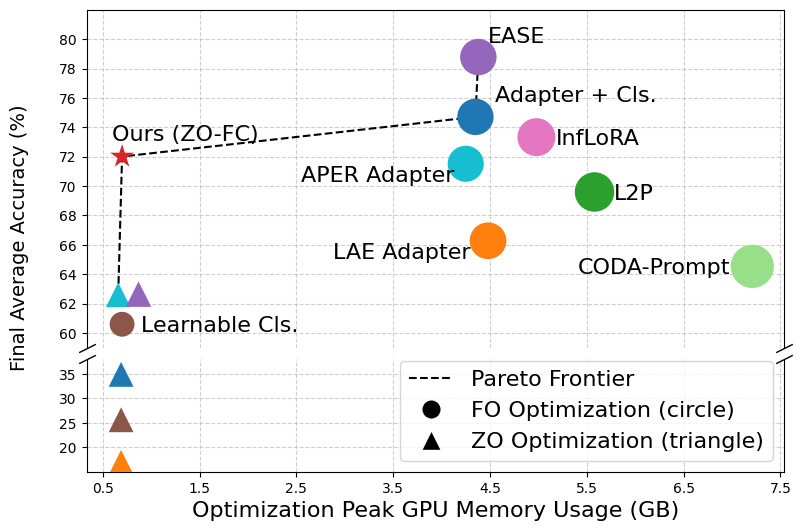}           
  \caption{Memory–accuracy trade-off across CIL methods on ImageNet-R. Note that, same color denotes the same method.}
  \label{fig:tradeoff_inr}          
\end{figure}

\noindent\textbf{Trade-off between accuracy and resource cost: } For on-device CL, peak training memory is often the hard bottleneck. As illustrated in Figure \ref{fig:tradeoff_inr}, ZO-FC demonstrates a better trade-off: 1) By leveraging the gradient-free nature of ZO optimization, ZO-FC requires only 0.7 GB of training memory, achieving up to a 6$\times$ reduction compared to FO methods. 2) ZO-FC significantly outperforms other ZO approaches in accuracy. We therefore foreground the accuracy-memory trade-off in the main text and provide compute and runtime details in supplementary material. Practically, the extra ZO forward overhead can be amortized via parallel or batched perturbation evaluation, e.g., P-RGE\cite{gao2024enabling} to support ZO-FC under real deployment.

\subsection{Analysis and Ablation Study}
\noindent\textbf{ZO-FC finds flatter shared region to mitigate forgetting: }
\label{sec: hybrid_flatness_measure}
To validate our analysis in Section \ref{sec:method}, which shows that ZO-FC guides the model toward wider and flatter minima that mitigate forgetting, we quantify loss-landscape flatness during CL. While the curvature of the loss landscape can be characterized by Hessian eigenvalues, it is computationally expensive to compute them directly. We adopt an approach inspired by Sharpness-Aware Minimization (SAM)~\cite{foret2021sharpnessaware} as a more efficient alternative, which perturbs model parameters within a small neighborhood and measures the induced loss change.
Concretely, for model parameters ($\theta$) and the old-task loss ($L_{\text{old}}$),
we define the flatness metric as:
\begin{equation}
  \Phi_{\mathrm{ZO}}(\theta;\rho)
  \;=\;
  \frac{L_{\mathrm{old}}(\theta^{+})
        -L_{\mathrm{old}}(\theta)}
       {\rho\,\bigl(L_{\mathrm{old}}(\theta)+\epsilon\bigr)},
  \label{eq:zo-sam}
\end{equation}
where $\theta^+$ represent the perturbed parameters with a small radius $\rho$, and $\epsilon = 10^{-12}$. The denominator is used to normalize the metric score. Lower score means flatter landscape.

\begin{figure}[!h]
  \centering
  \vspace{-0.2em}

  \begin{subfigure}[b]{\linewidth}
    \includegraphics[width=\linewidth]{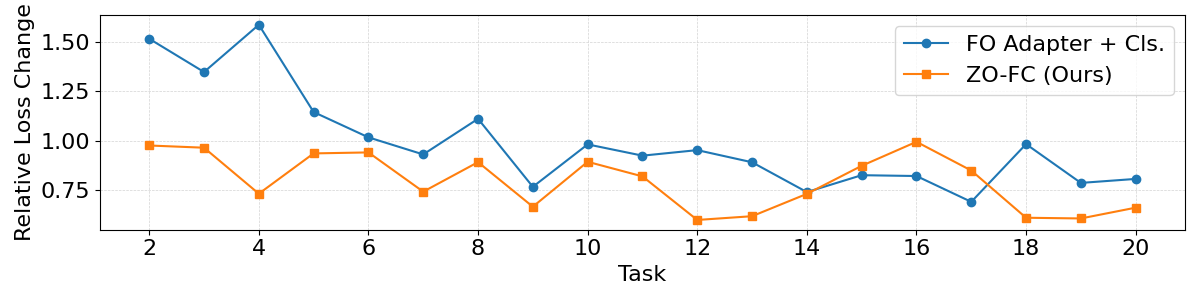}
    \caption{CIFAR100}
    \label{fig:zo_sam_cifar}
  \end{subfigure}
  \hfill
  \begin{subfigure}[b]{\linewidth}
    \includegraphics[width=\linewidth]{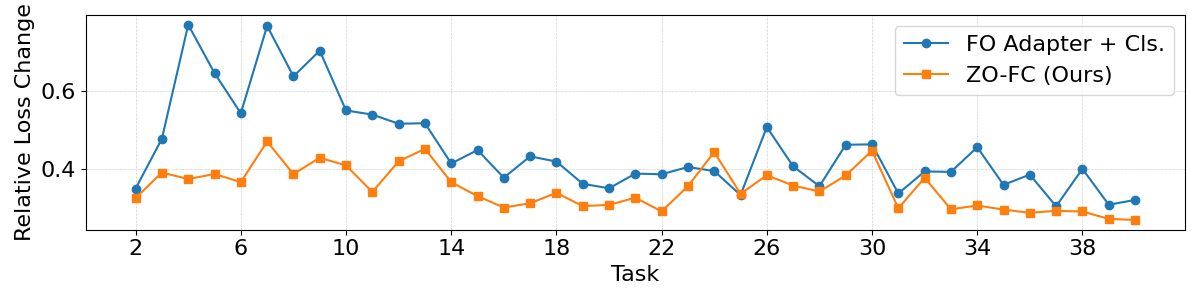}
    \caption{ImageNet-R}
    \label{fig:zo_sam_inr}
  \end{subfigure}
  \caption{SAM-based flatness score of FO and ZO optimization on CIFAR100 and ImageNet-R Inc-5.}
  \label{fig:zo_sam_flatness}
\end{figure}

As shown in Figure \ref{fig:zo_sam_flatness}, the SAM-based loss changes are consistently smaller under ZO-FC than FO. This indicates that ZO-FC tends to converge to flatter minima, matching its lower forgetting. The late-task ‘plateau’ comes from the proxy itself: with fixed $\rho$ and training already in a low-loss, near-quadratic regime, both the perturbation loss increase (numerator) and the current loss (denominator) shrink together; meanwhile averaging over more old tasks further smooths the curve. Thus the SAM curves narrow for all methods even while end-of-sequence metrics (last, fgt) still clearly favor ZO-FC. In short, ZO optimization on the adapter pushes features toward wider minima, while FO on the classifier supplies boundary plasticity.

\noindent\textbf{Ablation on the query number sensitivity in ZO-FC: } Table \ref{tab:hybrid query} varies the query number Q$\in${1,2,4,8,32,64}. On CIFAR100, accuracy changes little because coarse ZO estimates are sufficient for effective learning on such simpler dataset, making the benefit of additional queries marginal. On the more complex and more diverse ImageNet-R, accuracy rises with Q. As Q increases, the ZO gradient approximation becomes more accurate, improving model adaptation and overall accuracy, though the improvement grows slowly at higher Q when approaching saturation. Unlike accuracy, forgetting is not monotone with Q increasing. It can rise slightly at small–mid Q, then saturates or even drops once the estimator variance is low.


 \begin{table}[!h]
  \centering
  \small
  \setlength{\tabcolsep}{3pt}  
  \begin{tabular}{lcccccc}
    \toprule
    \multirow{2}{*}{Method} &
      \multicolumn{3}{c}{\textbf{CIFAR100 Inc-5}} &
      \multicolumn{3}{c}{\textbf{ImageNet-R Inc-5}} \\
    \cmidrule(lr){2-4}\cmidrule(lr){5-7}
      & avg$\uparrow$ & last$\uparrow$ & fgt$\downarrow$ & avg$\uparrow$ & last$\uparrow$ & fgt$\downarrow$ \\
    \midrule
    Q=1  & 87.9 & 82.36 & 4.61 & 65.31 & 60.27 & 3.0\\
    Q=2  & 88.15 & 82.69 & 4.72 & 69.17 & 64.05 & 4.0\\
    Q=4  & 88.39 & 83.34 & 5.26 & 72.01 & 66.63 & 4.40\\
    Q=8  & 88.18 & 82.96 & 5.83 & 72.72 & 67.15 & 4.20\\
    Q=16  & 88.54 & 83.45 & 6.0 & 73.30 & 67.13 & 5.38\\
    Q=32  & 88.82 & 84.09 & 6.03 & 73.65 & 68.13 & 4.97\\
    Q=64  & 88.79 & 84.11 & 6.08 & 73.98 & 68.17 & 4.91\\
    \bottomrule
  \end{tabular}
  \caption{Ablation study on the query number Q in ZO-FC.}
  \label{tab:hybrid query}
\end{table}

To fully match the strong FO CL methods (e.g., InfLoRA, EASE), ZO methods may require very large Q or more advanced ZO optimization strategies, which we leave for future work. Our primary focus is to show that \textit{under limited training budgets, ZO-FC achieves accuracy on par with many FO methods (e.g., L2P, LAE, APER, CODA-Prompt), clearly outperforms their direct ZO variants, and provides substantial training memory savings, making it highly promising for on-device continual learning in resource-constrained settings.}

\section{Conclusion}
\label{sec:conclusion}
In this work, we investigate the use of zeroth-order (ZO) optimization for continual learning. We begin with a theoretical analysis showing that ZO optimization inherently helps mitigate forgetting while suffering from less plasticity. To assess its practical viability, we conduct a comprehensive empirical study applying ZO optimization to existing PEFT-based CL methods and find that naively replacing first-order (FO) optimization with ZO optimization consistently fails, primarily due to the instability caused by applying ZO to the learnable classifier. Motivated by that, we propose ZO-FC, a simple yet effective approach that applies ZO optimization to the adapter with FO updates for the classifier. Experimental results demonstrate that ZO-FC achieves accuracy comparable to FO methods while significantly reducing training memory consumption. In future work, we aim to validate ZO-FC on harder fine-grained CL settings, demonstrate on-device training and explore parallelizing, low-variance ZO methods.


\newpage

\clearpage
\section*{Acknowledgments}
This work was supported in part by the U.S. National Science Foundation under Grant No.2348376. 
{\small
\bibliographystyle{ieeenat_fullname}
\bibliography{main}
}
\clearpage

\section*{Supplementary Material}
\label{sec:A_appendix}

\subsection*{A. Training, Evaluation Configurations}
\noindent\textbf{Training configurations: }
For our proposed ZO-FC method, we optimize the classifier using FO SGD with a learning rate of 0.01 that follows a cosine decay schedule; the adapter parameters are updated using the SPSA and ZO SGD with a constant learning rate of 0.01. For SPSA, we set the perturbation magnitude to $\varepsilon=0.001$, and the query budget to $Q=4$, and apply L2 norm clipping at a threshold of 1.0 to the estimated gradients for stability.  The training budgets are set per dataset: (1) On CIFAR100, the FO classifier is trained for 10 epochs, and the ZO adapter is trained for 20 epochs; (2) On ImageNet-R, the FO classifier is trained for 20 epochs, and the ZO adapter is trained for 40 epochs. For all baseline methods, we adopt the official implementations and follow the hyperparameter settings recommended in their original papers and the LAMDA-PILOT benchmark \cite{sun2025pilot}, running on 1$\times$A6000, to ensure a fair comparison. Specifically, for the baseline combining an adapter with a cosine classifier, we train for a unified 10 epochs on CIFAR100 and 20 epochs on ImageNet-R. All runs use a batch size of 48 to ensure fair comparison of memory usage during optimization.

\noindent\textbf{Evaluation Metrics: }
Let $K$ be the number of tasks. Let $A_{i,j}$ denote the test accuracy on task $j$ after completing training on task $i$ (so the final accuracies are $\{A_{K,j}\}_{j=1}^K$).
\noindent\textbf{Average accuracy (avg).}
\[
\mathrm{avg} \;=\; \frac{1}{K}\sum_{j=1}^{K} A_{K,j}.
\]
\noindent\textbf{Last-task accuracy (last).}
\[
\mathrm{last} \;=\; A_{K,K}.
\]
\noindent\textbf{Catastrophic forgetting (fgt): }
For each old task $j\!<\!K$, let $A_j^\star$ be the best accuracy that task ever achieved during training:
\[
A_j^\star \;=\; \max_{i \in \{j,\dots,K\}} A_{i,j}.
\]
Forgetting is the average drop from this best value to the final value:
\[
\mathrm{fgt} \;=\; \frac{1}{K-1}\sum_{j=1}^{K-1}\!\left( A_j^\star \;-\; A_{K,j} \right).
\]

\subsection*{B. Algorithm}
As shown in Algorithm \ref{alg:hybrid_alg}, ZO-FC updates the classifier $\psi$ with standard FO SGD while updating the PEFT adapter $\phi$ via a two-point ZO (SPSA) step. For each mini-batch, we first compute one FO loss and back-propagate \emph{only} through the head (Lines~1--3). We then draw $Q$ Rademacher directions $\{\Delta_q\}_{q=1}^Q$ and evaluate the loss at $\phi\pm\varepsilon\Delta_q$ (Lines~5--10), yielding the finite-difference estimates
$\hat g_q=\frac{L^+-L^-}{2\varepsilon}\,\Delta_q$ (Line~11). The ZO estimates are averaged and clipped before a ZO-SGD update on $\phi$ (Lines~12--14). 

\begin{algorithm}[h]
\caption{The proposed ZO-FC}
\begin{algorithmic}[1]
\REQUIRE mini batch $\mathcal{B}$, classifier $\psi$, adapter $\phi$, learning rates $\eta_\psi$, $\eta_\phi$; perturbation scale $\varepsilon$, query budget $Q$
\STATE \textbf{// FO step on classifier}
\STATE $L \gets L(\phi,\psi;\mathcal{B})$ \COMMENT{back-prop only through $\psi$}
\STATE $\psi \leftarrow \psi - \eta_\psi \nabla_\psi L$
\STATE \textbf{// ZO step on adapter (SPSA)}
\FOR{$q = 1$ to $Q$}
  \STATE sample $\Delta_q \sim \{\pm1\}$ \COMMENT{store $\Delta_q$ only}
  \STATE $\phi' \gets \phi + \varepsilon \Delta_q$
  \STATE $L^+ \gets L(\phi', \psi; \mathcal{B})$
  \STATE $L^- \gets L(\phi' - 2\varepsilon \Delta_q, \psi; \mathcal{B})$
  \STATE restore $\phi' \gets \phi$ \COMMENT{no need to store $\phi'$}
  \STATE $\hat{g}_q \gets \frac{L^+ - L^-}{2\varepsilon} \Delta_q$
\ENDFOR
\STATE $\bar{g}_\phi \gets \frac{1}{Q} \sum_q \hat{g}_q$
\STATE $\phi \leftarrow \phi - \eta_\phi\, \text{clip}(\bar{g}_\phi)$
\end{algorithmic}
\label{alg:hybrid_alg}
\end{algorithm}

\subsection*{C. Proof Assumptions for ZO SPSA searches flat minima}
Under the standard assumptions that $L$ is convex, three-times continuously differentiable, and its third derivatives are bounded (the usual ``$L_3$-smoothness'' condition), \citet{zhang2025zofindsflat} proved that in \(T=\tilde{\mathcal{O}}(d^{4}/\varepsilon^{2})\) iterations: SPSA drives the $\varepsilon$-smoothed loss \(\tilde{L}_{\varepsilon}(\theta)\) to within $\varepsilon$ of its global minimum: 
\begin{equation}
\tilde{L}_{\varepsilon}(\theta_T) - \min_{\theta} \tilde{L}_{\varepsilon}(\theta) \;\leq\; \varepsilon,
\label{eq:spsa-smooth-minima}
\end{equation}
where $\theta_T$ is the final parameters after $T$ SPSA steps.
By the definition of $\tilde{L}_{\varepsilon}$, \citet{zhang2025zofindsflat} also showed
\begin{equation}
\text{Tr} \left[ \nabla^2 \mathcal{L}(\theta_T) \right] - \min_{\theta^*} \text{Tr} \left[ \nabla^2 \mathcal{L}(\theta^*) \right]  \;\leq\; \varepsilon,
\label{eq:spsa-minima-bound}
\end{equation}
that is, among all minimizers of the original loss $L(\theta)$, SPSA selects those with the smallest Hessian trace-flatter minima.

Standard GD optimizes the raw loss $f$, which does not guarantee bias toward flatter minima. Empirically, SGD sometimes finds wide valleys due to noise, but this is incidental and depends on schedule and noise scale. Explicit methods such as \textbf{Entropy-SGD} \cite{chaudhari2019entropy}, \textbf{Sharpness-Aware Minimization (SAM)} \cite{foret2021sharpnessaware} bias GD trajectories into wide valleys by injecting perturbations into the update rule. In contrast, \textbf{two-point ZO} \textit{always} optimizes the smoothed loss $\tilde{f}$, and theory shows it minimizes to within $\epsilon$ of the smallest-Hessian-trace solution. This explains why our ZO-FC retains low forgetting across various streams: it consistently biases the shared adapter toward flat representations, without extra regularization.

\subsection*{D. Proof for flat minima in Training dynamics of ZO-FC}
When combining zeroth order adapter with first order classifier, the flat minima effect is maintained for continual learning. For the adapter $\phi$ and classifier $\psi$, a two-point SPSA step on $\phi$ is an unbiased estimator of the gradient of the \emph{Gaussian-smoothed} loss 
$\tilde{L}_{\varepsilon}(\phi,\psi)=\mathbb{E}_{u}[L(\phi+\varepsilon u,\psi)]$:
$\mathbb{E}[g_{\phi}]=\nabla_{\phi}\tilde{L}_{\varepsilon}(\phi,\psi)$.
If the classifier $\psi$ is updated on a faster time-scale (FO) while $\phi$ evolves slowly (ZO), the limiting dynamics follow
$\bar{\psi}=-\nabla_{\psi} L(\phi,\psi)$ (fast) with $\psi^\star(\phi)=\arg\min_{\psi}L(\phi,\psi)$, and 
$\bar{\phi}=-\nabla_{\phi}\tilde{L}_{\varepsilon}(\phi,\psi^\star(\phi))$ (slow). 
Hence the hybrid iterates descend the reduced objective
$J(\phi)=\min_{\psi}\tilde{L}_{\varepsilon}(\phi,\psi)$.
This connects ZO-FC to flat minima: smoothing acts directly on the representation $\phi$, biasing toward lower curvature for stability, while FO keeping the classifier plastic.

\subsection*{E. Component Analysis: Classifier Variants and FO, ZO swap}
We further investigate the design of our approach by comparing the impact of different learnable classifiers (cosine vs. linear) and trainable component optimization strategies (FO vs. ZO). Table \ref{tab:ablation_simple_cosine_linear_comparison} and Table \ref{tab:ZO components} present detailed results on the Inc-5 setting for both CIFAR100 and ImageNet-R.

\noindent\textbf{Classifier Choice: }
As shown in Table \ref{tab:ablation_simple_cosine_linear_comparison}, the FO optimized adapter performs closely no matter combined with either a cosine or linear classifier though cosine classifier works better than linear classifier when without adapter. FO-optimized adapters achieve strong performance regardless of whether they are paired with a cosine or linear classifier. However, the cosine classifier consistently yields higher accuracy and lower forgetting than its linear counterpart when used alone or within hybrid settings. When combined with linear classifiers, ZO adapter struggle with catastrophic forgetting, suggesting poor alignment between the decision boundary and the adapted feature space under noisy updates. This performance gap indicates that the cosine classifier better captures and reflects the fine-grained adaptations introduced by the ZO-optimized adapter, leading to superior retention and generalization.We also evaluate an EASE-style variant: the temporary FO classifier used during training is discarded and replaced by per-class prototypes after each task as final classifier. Replacing the temporary FO learnable classifier during training with prototypes as final classifier yields performance near SimpleCIL but clearly below ZO-FC and with higher forgetting.

\begin{table}[h]
  \centering
  \small
  \setlength{\tabcolsep}{2.5pt}  
  \begin{tabular}{lcccccc}
    \toprule
    \multirow{2}{*}{Method} &
      \multicolumn{3}{c}{\textbf{CIFAR100 Inc-5}} &
      \multicolumn{3}{c}{\textbf{ImageNet-R Inc-5}} \\
    \cmidrule(lr){2-4}\cmidrule(lr){5-7}
      & avg & last & fgt & avg & last & fgt \\
    \midrule
    Cosine Cls. & 85.61 & 78.33 & 5.17 & 60.6 & 52.12 & 3.98 \\
    Linear Cls. & 84.78 & 76.24 & 9.32 & 61.19 & 44.42 & 17.63 \\
    Adapter + Cosine Cls. & 88.41 & 83.64 &  9.58 & 74.71 & 69.62 & 8.49 \\
    Adapter + Linear Cls.  & 90.19 & 83.97 & 7.69 & 73.08 & 66.12 & 9.01  \\
    ZO-FC (Cosine)   & 88.39 & 83.34 & 5.26 & 72.01 & 66.63 & 4.40 \\
    ZO-FC (Linear)  & 85.94 & 79.61 & 5.92 & 62.07 & 51.78 & 9.89  \\
    ZO Adapter + Prototype & 84.94 & 80.87 & 10.77 & 67.06 & 62.98 & 16.39\\
    \bottomrule
  \end{tabular}
  \caption{Ablation study of classifier types under FO optimization of the classifier and ZO updates on the adapter}
  \label{tab:ablation_simple_cosine_linear_comparison}
\end{table}

\noindent\textbf{Effect of ZO Placement: }
Table \ref{tab:ZO components} isolates the impact of applying ZO optimization to different trainable components of the model. When the adapter is optimized with ZO and the classifier remains FO, that is, ZO-FC, the model achieves strong performance. In contrast, reversing this setup and applying ZO to the classifier while keeping the adapter FO optimized results in remarkable performance degradation. This contrast confirms that the classifier requires precise gradient-based optimization to maintain stable decision boundaries across tasks, while the adapter can tolerate noisy ZO updates without compromising overall performance.

\begin{table}[h]
  \centering
  \small
  \setlength{\tabcolsep}{3pt}  
  \begin{tabular}{lcccccc}
    \toprule
    \multirow{2}{*}{Method} &
      \multicolumn{3}{c}{\textbf{CIFAR100 Inc-5}} &
      \multicolumn{3}{c}{\textbf{ImageNet-R Inc-5}} \\
    \cmidrule(lr){2-4}\cmidrule(lr){5-7}
      & avg & last & fgt & avg & last & fgt \\
    \midrule
    ZO-FC  & 88.39 & 83.34 & 5.26 & 72.01 & 66.63 & 4.40\\
    FO adapter + ZO Cls. & 62.45 & 53.22 & 25.65 & 23.35 & 11.72 & 18.07 \\
    \bottomrule
  \end{tabular}
  \caption{Effect of ZO,FO components placement}
  \label{tab:ZO components}
\end{table}

These results separate the roles of representation smoothing (adapter) and decision-boundary plasticity (classifier), and motivate our choice of a single shared adapter.

\subsection*{F. Data Augmentation and Effects}
All image augmentations are applied in the dataloader before the model forward. No augmentation is performed inside the forward pass. This pipeline is identical for FO, ZO, and ZO-FC. Stochastic layers in the backbone (e.g., dropout) are disabled during ZO updates. The detailed augmentations are shown below, following configurations in \cite{sun2025pilot} and \cite{liang2024inflora}:\\
\noindent\textbf{CIFAR100, ImageNet-R.} \emph{Train:} RandomResizedCrop( 224, scale $\in$(0.05,1.0), ratio $\in$(3/4,4/3) ), RandomHorizontalFlip( p=0.5 ). \emph{Test:} Resize(256), CenterCrop(224).\\
\noindent\textbf{DomainNet.} \emph{Train:} RandomResizedCrop(224), RandomHorizontalFlip, Normalize($\mu$=(0,0,0),$\sigma$=(1,1,1)). \emph{Test:} Resize(256), CenterCrop(224), Normalize($\mu$=(0,0,0), $\sigma$=(1,1,1)).

 \begin{table}[h]
  \centering
  \small
  \setlength{\tabcolsep}{3pt}  
  \begin{tabular}{lcccccc}
    \toprule
    \multirow{2}{*}{Method} &
      \multicolumn{3}{c}{\textbf{CIFAR100 Inc-5}} &
      \multicolumn{3}{c}{\textbf{ImageNet-R Inc-5}} \\
    \cmidrule(lr){2-4}\cmidrule(lr){5-7}
      & avg & last & fgt & avg & last & fgt \\
    \midrule
    ZO-FC w aug  & 88.39 & 83.34 & 5.26 & 72.01 & 66.63 & 4.40\\
    ZO-FC w/o aug & 88.63 & 83.14 & 8.02 & 73.04 & 67.35 & 5.19\\ 
    \bottomrule
  \end{tabular}
  \caption{Effect of data augmentation on ZO-FC.}
  \label{tab:hybrid query}
\end{table}

Augmentations act as input-space regularization that encourages invariances. In continual learning, that regularization reduces representation drift toward the current task and thus mitigates forgetting. On CIFAR100, removing augmentation (keeping only resize/center-crop) yields similar accuracy, and higher forgetting; on ImageNet-R, the accuracy increases slightly with much higher forgetting. This supports that standard augmentation helps ZO-FC stability rather than hurting ZO quality.

\subsection*{G. Further studies on Prompt-based Methods}
\begin{table}[h]
  \centering
  \small
  \setlength{\tabcolsep}{3pt}  
  \begin{tabular}{lcccccc}
    \toprule
    \multirow{2}{*}{Method} &
      \multicolumn{3}{c}{\textbf{CIFAR100 Inc-5}} &
      \multicolumn{3}{c}{\textbf{ImageNet-R Inc-5}} \\
    \cmidrule(lr){2-4}\cmidrule(lr){5-7}
      & avg & last & fgt & avg & last & fgt \\
    \midrule
    L2P  & 26.43 & 4.79 & 19.05 & 11.83 & 2.52 & 10.21\\
    CODA-Prompt & 10.12 & 2.40 & 11.53 & 6.47 & 1.21 & 5.83\\ 
    \bottomrule
  \end{tabular}
  \caption{Increased epochs for prompt based method, 100 for Cifar100, 200 for INR.}
  \label{tab:increased zo prompt epoch}
\end{table}
We conduct further experiments on L2P and CODA-Prompt with more updates (100 epochs on CIFAR100, 200 on ImageNet-R). Despite the longer schedule, the two prompt based methods fail to converge under ZO optimization and their accuracy results remain far below adapter-based ZO training and ZO-FC. This matches recent studies \cite{yang2024adazeta} that prompt-based methods are harder to converge under ZO SPSA because optimizing prompt selection with penalties and contrastive alignment loss requires precise gradients.

\subsection*{H. Memory and Computation Cost Analysis}
We evaluate the memory efficiency and computational overhead of our ZO-FC method relative to both FO and ZO optimized CIL methods. This is critical for continual learning in resource-constrained environments, such as edge devices.

\begin{figure}[h]              
  \centering
  \includegraphics[width=1.0\linewidth]{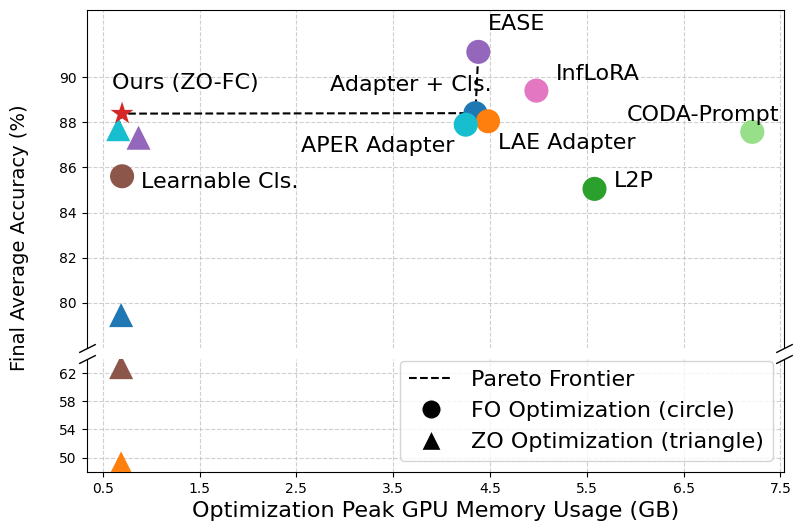}           
  \caption{Memory–accuracy trade-off across CIL methods on
CIFAR100. Note that, same color denotes the same method.}
  \label{fig:tradeoff_cifar}          
\end{figure}

\begin{figure}[h]              
  \centering
  \includegraphics[width=1.0\linewidth]{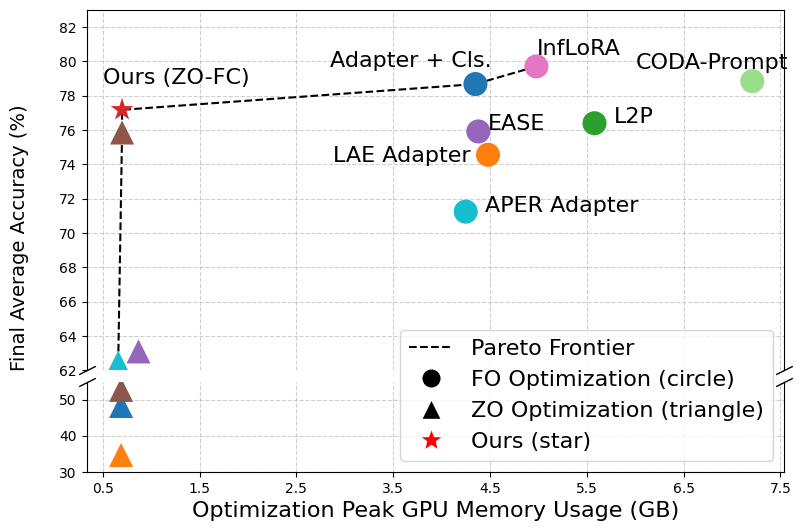}        
  \caption{Memory–accuracy trade-off across CIL methods on
DomainNet. Note that, same color denotes the same method.}
  \label{fig:tradeoff_domainnet}          
\end{figure}

\noindent\textbf{Memory–Accuracy Trade-off: }
As shown in Table \ref{tab:peak_mem}, FO optimization methods that incorporate PEFT modules incur significant memory costs—up to 7.21 GB for CODA-Prompt and over 4 GB for most adapter-based baselines. In contrast, ZO optimization yields dramatic memory savings, with most ZO variants requiring $< 0.7$ GB, regardless of the adapter type. Notably, our proposed ZO-FC maintains only 0.70 GB of peak GPU memory usage—on par with minimal FO models using fixed classifiers. This demonstrates that our hybrid strategy preserves the memory efficiency of ZO methods, while achieving high performance, as shown in Figure \ref{fig:tradeoff_cifar}, \ref{fig:tradeoff_domainnet}.

\begin{table}[h]
\centering
\small
\begin{tabular}{lc}
\toprule
Scenario & Peak GPU Memory (GB)\\
\midrule
\midrule
\textbf{FO Optimization}\\
\midrule
Learnable Cls. & 0.70 \\
Adapter + Learnable Cls.  & 4.35   \\
LAE  &  4.48    \\
L2P  &   5.58  \\
CODA-Prompt  &  7.21   \\
APER  &  4.25  \\
EASE  &  4.38  \\
InfLoRA & 4.98 \\
\midrule
\textbf{ZO Optimization}\\
\midrule
Learnable Cls. & 0.69 \\
Adapter + Learnable Cls.  &  0.69    \\
LAE & 0.69 \\
APER & 0.66\\
EASE & 0.87\\
InfLoRA & 0.88 \\
\midrule
ZO-FC & 0.70   \\
\bottomrule
\end{tabular}
\caption{Per epoch Optimization Memory cost with batch size 48}
\label{tab:peak_mem}
\end{table}

\begin{table}[h]
\centering
\small
\begin{tabular}{lcc}
\toprule
Scenario & Peak GPU Memory Cost & Running time \\
\midrule
FO-SGD  & 4.35 GB  & 24.54s \\
ZO-SGD (q=1) &  0.69 GB   &  19.44s\\
ZO-SGD (q=4) &  0.69 GB   & 52.31s \\
ZO-FC (q=1) & 0.70 GB  & 22.15s \\
ZO-FC (q=4) & 0.70 GB  & 59.36s \\
\bottomrule
\end{tabular}
\caption{Per epoch Memory cost \& Time Cost with batch size 48}
\label{tab:Runtime}
\end{table}

\noindent\textbf{Runtime–Memory Trade-off: }
Table \ref{tab:Runtime} shows the per-epoch training time and corresponding optimization memory cost. ZO methods achieve extreme memory efficiency, but incur increasing compute overhead with higher query budgets, reflecting the cost of additional function queries. Due to optimizer decomposition for different trainable modules, our approach costs slightly longer runtime than ZO. Despite this trade-off, Our approach's runtime remains manageable to low-resource hardware.

\noindent\textbf{Computation Cost Analysis: }
We use FLOPs to measure computation cost. Let \textbf{B} be batch size; \textbf{H}$\times$\textbf{W} the input resolution ($224{\times}224$);
\textbf{N} the ViT token count ($N = 1 + (H/16)(W/16) = 197$ at $224^2$); \textbf{D} the hidden width (ViT-B: $D{=}768$);
\textbf{C} the current \#classes; $\mathbf{L_a}$ the \#blocks with adapters; and \textbf{r} the adapter bottleneck (we use $r{=}5$).

Denote the ViT backbone forward cost per image as $F_{\text{backbone}}(H,W)$.
A bottleneck adapter applies two linear maps per token ($D{\to}r$ and $r{\to}D$), so the adapter overhead per image is
\begin{equation}
F_{\text{PET}} \;\approx\; L_a \, N \, (2Dr + 2rD) \;=\; L_a\,N\,(4Dr).
\label{eq:flops_pet}
\end{equation}

For the \emph{cosine} classifier, each forward computes: feature $\ell_2$–norm ($\sim 3D$ ops),
\emph{weight} $\ell_2$–norm across $C$ classes ($\sim 3CD$ ops), class dot-products ($\sim 2CD$), and a scalar scale $\sigma$ ($\sim C$):
\begin{equation}
F_{\text{cls}} \;\approx\; (5CD) + 3D + C.
\label{eq:flops_cls}
\end{equation}
\textit{Note.} In our implementation, feature and weight normalizations are recomputed \emph{every forward}, so the $3CD$ term in \eqref{eq:flops_cls} is paid per forward. If one caches unit-norm weights (or renormalizes once per optimizer step), then the per-forward cost becomes $(2CD)+3D+C$ with an additional $3CD$ once per step.

The total \emph{per-image forward} FLOPs is
\begin{equation}
F_{\text{fwd,img}} \;\approx\; F_{\text{backbone}}(H,W) \;+\; L_a N (4Dr) \;+\; \big(5CD + 3D + C\big).
\label{eq:flops_img_total}
\end{equation}

For per-batch multiplicities (FO / ZO / hybrid), We report FLOPs using the standard convention that a backward pass over the trainable subgraph costs $\approx$2× the forward pass, so one training step is $\approx$3× the forward FLOPs. So let $\alpha_{\text{FC}}$ be the FC-classifier backward cost in forward-equivalents of the \emph{head} subgraph (we use $\alpha_{\text{FC}}\!\approx\!2$), and let $q$ be the ZO (SPSA) query budget:
\begin{itemize}
  \item \textbf{FO:} forwards/batch $=1$; backward over \emph{trainable parts} (adapters+head). Cost: $\;B\,F_{\text{fwd,img}} \;+\; \alpha_{\text{FC}}\,B\,F_{\text{cls}}$ plus the adapter backward (captured by the 3$\times$ rule below).
  \item \textbf{ZO on adapters:} forwards/batch $= 2q{+}1$; no backward. Cost: $\;(2q{+}1)\,B\,F_{\text{fwd,img}}$.
  \item \textbf{ZO-FC (early epochs):} $2q$ ZO forwards on adapters $+$ one FO forward for the classifier $+$ one eval forward; and a classifier-only backward. Cost: $\;(2q{+}2)\,B\,F_{\text{fwd,img}} \;+\; \alpha_{\text{FC}}\,B\,F_{\text{cls}}$.
  \item \textbf{ZO-FC (late; classifier frozen):} same forwards as ZO: $2q{+}1$. Cost: $\;(2q{+}1)\,B\,F_{\text{fwd,img}}$.
\end{itemize}

Using \eqref{eq:flops_img_total}, the relative \emph{per-batch} overhead vs.\ FO is
\begin{align}
\Delta F_{\text{ZO}\;\text{vs}\;\text{FO}} &\approx 2q \cdot B \cdot F_{\text{fwd,img}},\\
\Delta F_{\text{ZO-FC(early)}\;\text{vs}\;\text{FO}} &\approx (2q{+}1) \cdot B \cdot F_{\text{fwd,img}},\\
\Delta F_{\text{ZO-FC(late)}\;\text{vs}\;\text{FO}} &\approx 2q \cdot B \cdot F_{\text{fwd,img}}.
\end{align}
\textit{Remarks.} (i) In FO we backprop through the \emph{trainable} parts (adapters+classifier); the ViT backbone is frozen (no grads) but still incurs one forward per batch.
(ii) For ViT-B/16 at $224^2$, $N{=}197$ and $D{=}768$; the adapter overhead scales linearly with $r$ and is small for $L_a{=}5$, $r{=}5$.
\begin{table*}[h]
\centering
\small
\setlength{\tabcolsep}{7pt}
\renewcommand{\arraystretch}{1.16}
\caption{Per-batch FLOP \emph{breakdown} for FO, ZO, and ZO-FC. 
\textit{Forward FLOPs / batch} are \textbf{measured} once with \texttt{fvcore} and scaled by the number of forwards in each regime (baseline single-forward cost: $8.484\times10^{11}$). 
\textit{Backward FLOPs / batch} are \textbf{estimated}: FO via the $3\times$ rule; ZO-FC via the analytical head-only term $2B\,F_{\text{cls}}$ with $B{=}48,\,D{=}768,\,C{=}5$ at the measured epoch.}
\label{tab:flops-breakdown}
\begin{tabular}{l c c c c c c}
\toprule
\textbf{Regime} & \textbf{q} & \textbf{Forwards / batch} &
\textbf{Forward FLOPs (meas.)} &
\textbf{Backward FLOPs (est.)} &
\textbf{Total FLOPs (est.)} & \(\times\)\textbf{FO} \\
\midrule
FO                          & -- & $1$            & $0.848\times10^{12}$ & $1.697\times10^{12}$ & $\mathbf{2.545\times10^{12}}$ & $\mathbf{1.00\times}$ \\
\midrule
ZO                          & 1  & $2q{+}1{=}3$   & $\mathbf{2.545\times10^{12}}$ & $0$                  & $\mathbf{2.545\times10^{12}}$ & $\mathbf{1.00\times}$ \\
ZO                          & 4  & $2q{+}1{=}9$   & $\mathbf{7.636\times10^{12}}$ & $0$                  & $\mathbf{7.636\times10^{12}}$ & $\mathbf{3.00\times}$ \\
\midrule
ZO\textnormal{-}FC (early)  & 1  & $\approx 2q{+}1$ & $\mathbf{2.545\times10^{12}}$ & $2.06\times10^{6}$    & $\mathbf{2.545\times10^{12}}$ & $\mathbf{1.00\times}$ \\
ZO\textnormal{-}FC (late)   & 1  & $2q{+}1{=}3$   & $\mathbf{2.545\times10^{12}}$ & $0$                  & $\mathbf{2.545\times10^{12}}$ & $\mathbf{1.00\times}$ \\
ZO\textnormal{-}FC (early)  & 4  & $\approx 2q{+}1$ & $\mathbf{7.636\times10^{12}}$ & $2.06\times10^{6}$    & $\mathbf{7.636\times10^{12}}$ & $\mathbf{3.00\times}$ \\
ZO\textnormal{-}FC (late)   & 4  & $2q{+}1{=}9$   & $\mathbf{7.636\times10^{12}}$ & $0$                  & $\mathbf{7.636\times10^{12}}$ & $\mathbf{3.00\times}$ \\
\bottomrule
\end{tabular}
\raggedright\footnotesize 
Notes: (i) FO backward is estimated via the rule-of-thumb \(\text{FLOPs}_{\text{total}}\!\approx\!3\times\text{FLOPs}_{\text{forward}}\) when training adapters+classifier with a frozen backbone. 
(ii) ZO-FC early includes a classifier-only FO step; its backward is $2B\,F_{\text{cls}}$ and scales linearly with $C$. 
(iii) Because this term is tiny relative to ViT forwards, totals for ZO and ZO-FC practically coincide at the same $q$. \par
\end{table*}

\noindent\textbf{Computation Cost Measurement: }
We use ViT-B/16 at $224^2$ with batch size $B{=}48$. The \emph{measured} forward FLOPs per image (fvcore) is $\mathbf{1.768\times 10^{10}}$,
so per-batch forward FLOPs is $\mathbf{8.484\times 10^{11}}$. FO total is \emph{estimated} via the common rule
$\text{FLOPs}_{\text{FO,total}}\!\approx\!3\times\text{FLOPs}_{\text{forward}}$, yielding $\mathbf{2.545\times 10^{12}}$ per batch.
ZO performs $(2q{+}1)$ forward evaluations with no backward; at $q{=}1$ this matches FO’s total, and at $q{=}4$ it is $9\times$ the forward-only batch cost, i.e., $\mathbf{7.636\times 10^{12}}$.
ZO-FC uses the same ZO forwards and adds a classifier-only backward during early epochs. Given our small $C$ and a frozen backbone, this extra term ($\alpha_{\text{FC}}\,B\,F_{\text{cls}}$) is orders of magnitude smaller than a ViT forward, so totals numerically coincide with ZO at the same $q$ (both early and late).
This explicitly reports and quantifies the \emph{additional forward passes} required by ZO and how training cost scales with $q$.
In practice, parallelizing the $2q$ ZO forwards~\cite{gao2024enabling} can reduce wall-clock time without changing FLOPs and is compatible with ZO-FC for on-device continual learning.

\subsection*{I. CIL Accuracy Trends: ZO-FC vs. FO Methods}
\begin{figure}[h]
  \centering
  \includegraphics[width=\columnwidth]{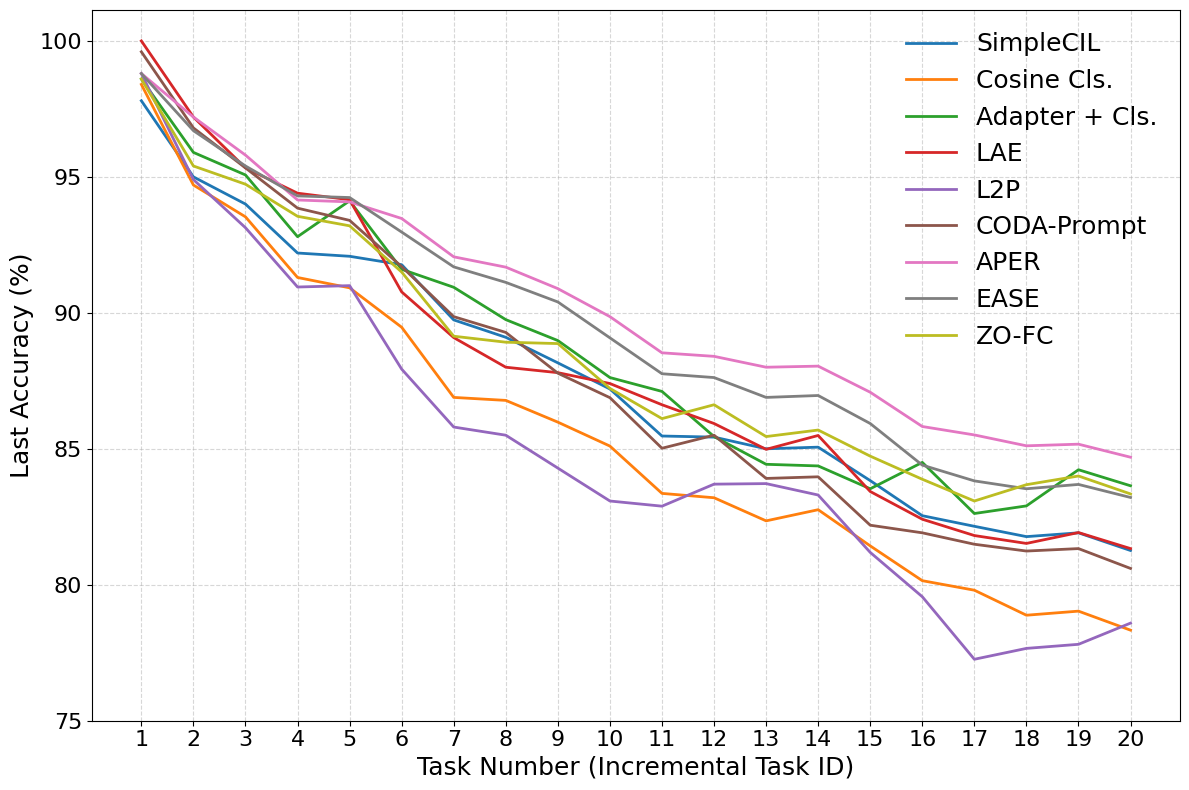}
  \caption{CIFAR100, Inc-5: last accuracy over tasks.}
  \label{fig:overview_inc5_cifar}
\end{figure}

\begin{figure}[h]
  \centering
  \includegraphics[width=\columnwidth]{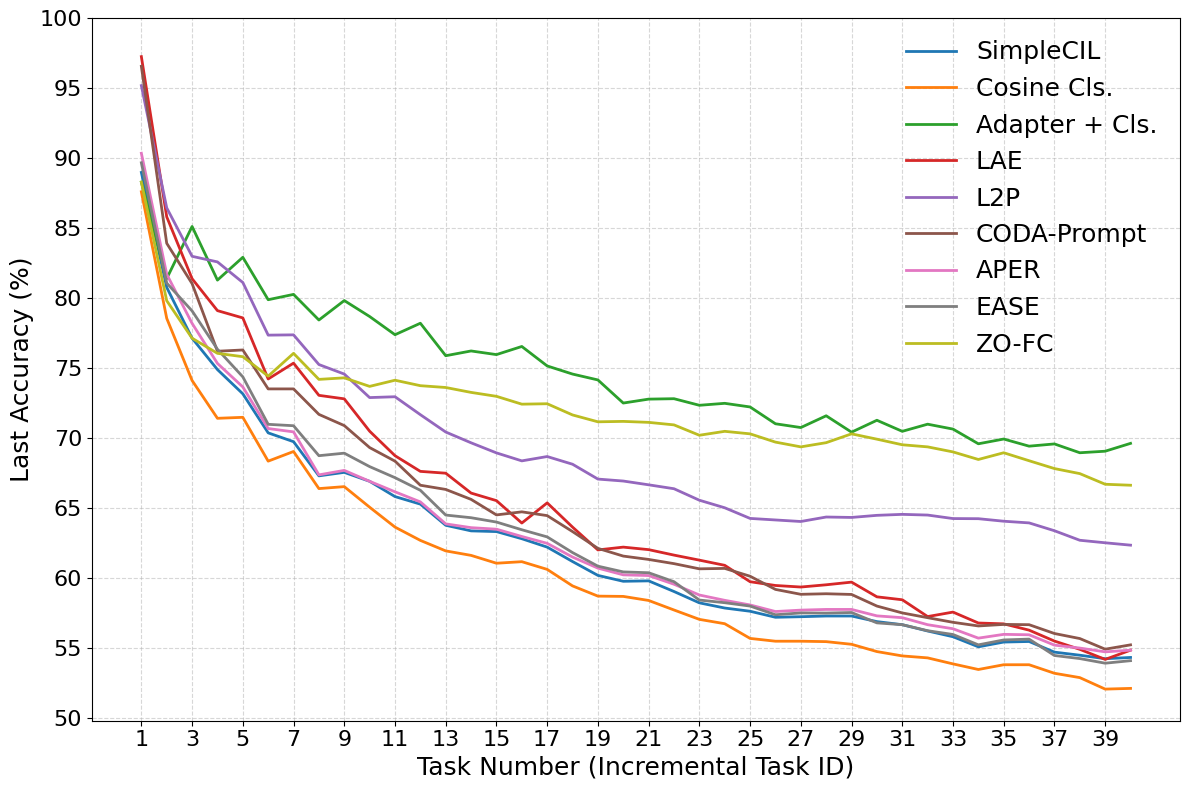}
  \caption{ImageNet-R, Inc-5: last accuracy over tasks.}
  \label{fig:overview_inc5_inr}
\end{figure}

\begin{figure}[h]
  \centering
  \includegraphics[width=\columnwidth]{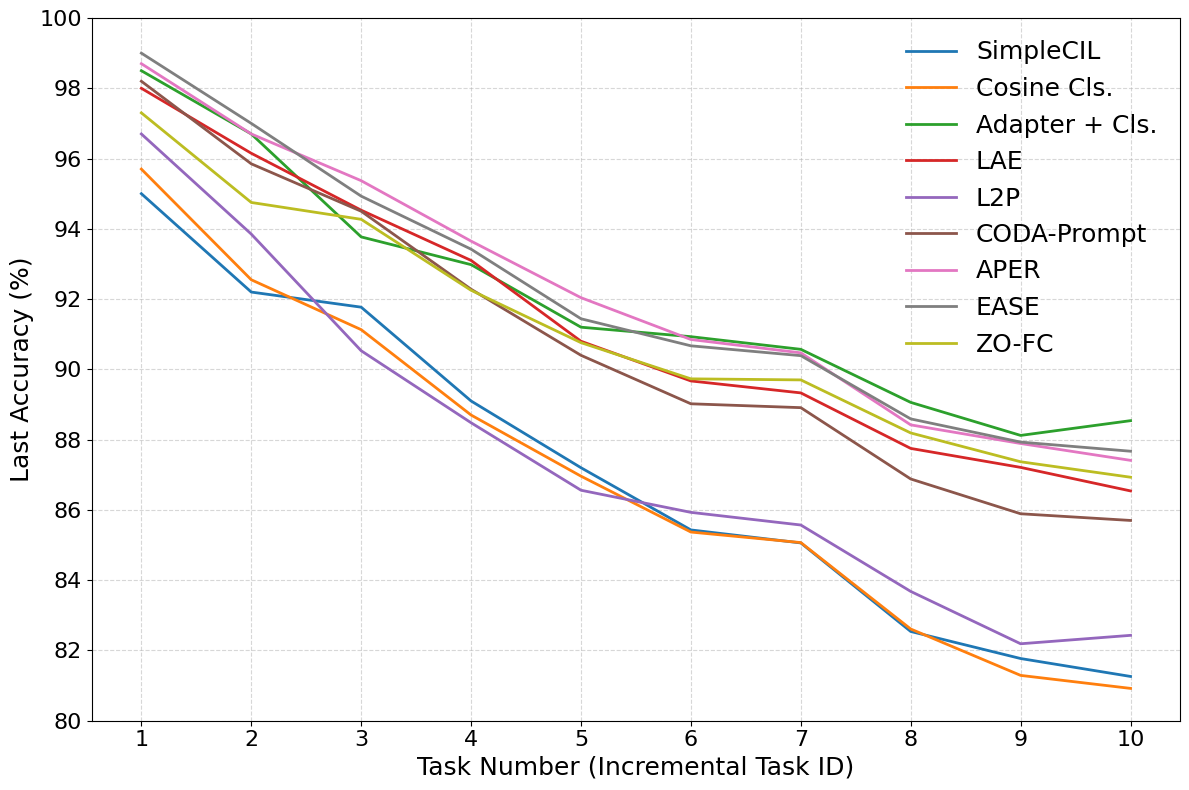}
  \caption{CIFAR100, Inc-10: last accuracy over tasks.}
  \label{fig:overview_inc10_cifar}
\end{figure}

\begin{figure}[h]
  \centering
  \includegraphics[width=\columnwidth]{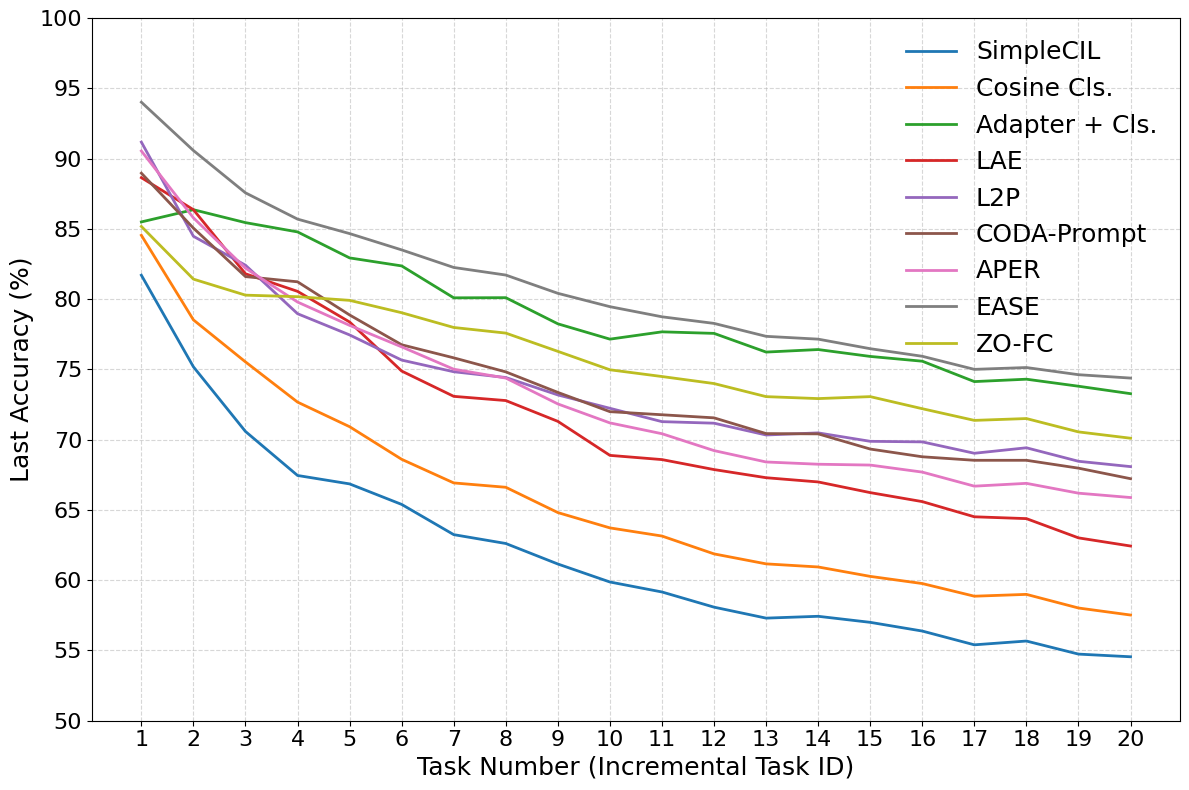}
  \caption{ImageNet-R, Inc-10: last accuracy over tasks.}
  \label{fig:overview_inc10_inr}
\end{figure}

To have a good view of the effectiveness of our proposed ZO-FC method, we demonstrate the last accuracy trends across the full task stream on both CIFAR100 and ImageNet-R, under two common incremental learning settings: Inc-10 and Inc-5. Figure \ref{fig:overview_inc5_cifar}, \ref{fig:overview_inc5_inr}, \ref{fig:overview_inc10_cifar}, \ref{fig:overview_inc10_inr} illustrate the evolution of performance as more tasks are introduced. These trends highlight the long-term stability of ZO-FC under class-incremental learning.

\end{document}